\documentclass{article} 
\usepackage{iclr2026_conference,times}

\usepackage{amsmath,amsfonts,bm}

\def\eqref#1{equation~\ref{#1}}

\def\1{\bm{1}}

\DeclareMathAlphabet{\mathsfit}{\encodingdefault}{\sfdefault}{m}{sl}
\SetMathAlphabet{\mathsfit}{bold}{\encodingdefault}{\sfdefault}{bx}{n}

\usepackage{hyperref}
\usepackage{url}
\usepackage{amsmath}   
\usepackage{amssymb}   
\usepackage{amsthm}    
\usepackage{booktabs}
\usepackage{makecell}
\usepackage{array}
\usepackage{longtable}
\usepackage{latexsym}
\usepackage{booktabs}
\usepackage{adjustbox}
\usepackage{multicol}
\usepackage{multirow}
\usepackage{pdflscape}
\usepackage{philex}
\usepackage{enumitem}
\usepackage{tabularx}
\usepackage{tcolorbox}
\usepackage{subcaption}  
\tcbuselibrary{listings, breakable}
\usepackage[table]{xcolor}

\title{Leveraging LLM Parametric Knowledge for Fact Checking without Retrieval}

\author{
 \textbf{Artem Vazhentsev\textsuperscript{*, 5}},
 \textbf{Maria Marina\textsuperscript{*, 2,1}},
 \textbf{Daniil Moskovskiy\textsuperscript{2,1}},
 \textbf{Sergey Pletenev\textsuperscript{2,1}},
 \\
 \textbf{Mikhail Seleznyov\textsuperscript{2,1}},
 \textbf{Mikhail Salnikov\textsuperscript{2,1}},
 \textbf{Elena Tutubalina\textsuperscript{2}},
 \textbf{Vasily Konovalov\textsuperscript{2,4}},
\\
 \textbf{Irina Nikishina\textsuperscript{3}},
 \textbf{Alexander Panchenko\textsuperscript{1,2}},
 \textbf{Viktor Moskvoretskii\textsuperscript{6}}
\\
 \textsuperscript{1}S-NLP Group,
 \textsuperscript{2}AXXX,
 \textsuperscript{3}Independent Researcher,
 \textsuperscript{4}MIRAI,
 \textsuperscript{5}MBZUAI,
 \textsuperscript{6}EPFL
\\
\texttt{
\href{mailto:Artem.Vazhentsev@mbzuai.ac.ae}{Artem.Vazhentsev@mbzuai.ac.ae}
  }
}

\iclrfinalcopy 
\begin{document}
\let\thefootnote\relax\footnotetext{* Equal contribution.}

\maketitle

\begin{abstract}
Trustworthiness is a core research challenge for agentic AI systems built on Large Language Models (LLMs). To enhance trust, natural language claims from diverse sources, including human-written text, web content, and model outputs, are commonly checked for factuality by retrieving external knowledge and using an LLM to verify the faithfulness of claims to the retrieved evidence. As a result, such methods are constrained by retrieval errors and external data availability, while leaving the model’s intrinsic fact-verification capabilities largely unused.
We propose the task of \textbf{fact-checking without retrieval}, focusing on the verification of arbitrary natural language claims, independent of their source. To study this setting, we introduce a comprehensive evaluation framework focused on generalization, testing robustness to (i) long-tail knowledge, (ii) variation in claim sources, (iii) multilinguality, and (iv) long-form generation. Across 9 datasets, 18 methods and 3 models, our experiments indicate that logit-based approaches often underperform compared to those that leverage internal model representations.
Building on this finding, we introduce \textbf{INTRA}, a method that exploits interactions between internal representations and achieves state-of-the-art performance with strong generalization. 
More broadly, our work establishes fact-checking without retrieval as a promising research direction that can complement retrieval-based frameworks, improve scalability, and enable the use of such systems as reward signals during training or as components integrated into the generation process.
\end{abstract}

\section{Introduction}

Ensuring the factual correctness of natural language claims, regardless of whether they originate from human-written text, web content, or automated systems, remains a fundamental challenge in the era of AI-generated texts.
One prominent source of incorrect claims is Large Language Models (LLMs), which are known to hallucinate and produce factually incorrect statements~\citep{DBLP:journals/corr/abs-2401-11817}. 
Hallucinations pose significant risks in high-stakes domains such as medicine, law, and science, making reliable detection a critical research priority~\citep{asgari2025clinical,DBLP:journals/nature/FarquharKKG24}.

Modern fact-checking methods predominantly rely on retrieval-based pipelines, such as FActScore~\citep{min-etal-2023-factscore} and SAFE~\citep{DBLP:conf/nips/WeiYSLH0TPLHDL24}. These approaches operate at the claim level: model outputs are first decomposed into \emph{atomic claims}, defined as minimal factual units that are objective and unambiguous. Each atomic claim is then verified by checking its faithfulness to retrieved external evidence, rather than assessing factual correctness in isolation.
While widely used, retrieval-based methods face several challenges: \begin{enumerate}[label=(\roman*), leftmargin=20pt, itemsep=0.3pt, topsep=0.1pt]
    \item They prioritize retrieved context over the factual knowledge encoded in an LLM’s parameters.
    \item They increase latency, as each generation requires querying external databases.
    \item Their effectiveness critically depends on retrieval quality, where noisy or irrelevant evidence can undermine the pipeline and lead to missed or false detections~\citep{DBLP:conf/sigir/CuconasuTSFCMTS24}.
\end{enumerate}
As a result, these detectors are constrained by the retrieval process and fail to fully exploit the model’s internal knowledge, limiting their scalability and robustness in practice.

By contrast, LLMs encode substantial factual knowledge in their parameters through large-scale pretraining and fine-tuning. Prior work shows that they store encyclopedic and commonsense facts, producing factually correct statements without external grounding~\citep{DBLP:journals/corr/abs-2010-11967,DBLP:journals/corr/abs-2207-05221}. Yet this intrinsic capability is rarely evaluated or exploited explicitly for hallucination detection.

\begin{figure}[t!]
    \centering
    \includegraphics[trim=0cm 0cm 0cm 0cm, width=\linewidth]{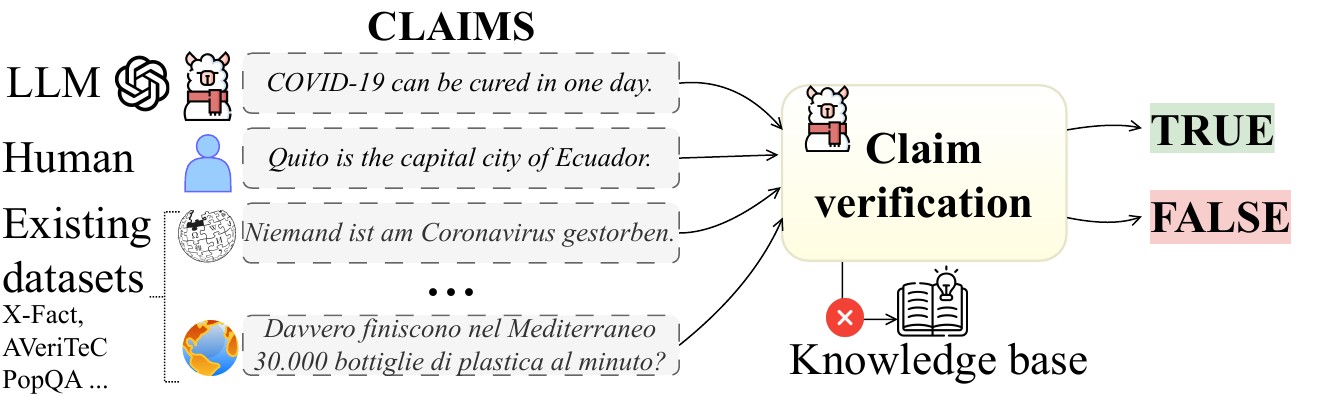}
    \caption{The task setting of fact-checking without retrieval. Claims from any source (human or LLMs) can be verified without having access to a knowledge base.}
    \label{fig:placeholder}

\end{figure}

In this work, we bridge this gap by introducing a new setting: \textbf{fact-checking without retrieval}. 
The task is to determine the factuality of an \emph{atomic claim} using only the LLM’s internal knowledge, without external retrieval. This setting is unique in two key ways:
\begin{enumerate}[label=(\roman*), leftmargin=20pt, itemsep=0.3pt, topsep=0.2pt]
    \item Unlike retrieval-based verifiers, it targets factual correctness rather than faithfulness to retrieved context.
    \item Unlike uncertainty estimation, it evaluates the factuality of arbitrary claims, rather than confidence over a model’s own generations.
\end{enumerate}

To investigate this setting, we introduce an evaluation framework focused on out-of-domain generalization, spanning 9 datasets that test (i) long-tail knowledge, (ii) claim source variation between human-made and generated claims, (iii) multilingual claims, (iv) long-form generation, and (v) cross-model claims. 
Across this framework, we evaluate 18 methods with 3 models and observe that detectors leveraging internal model representations consistently outperform logit-based uncertainty signals. 
Building on this observation, we propose \textbf{INTRA}, a simple yet effective internal-based method that leverages interactions between representations, achieving SoTA average performance and consistently strong generalization.

Our proposed setting opens several promising directions for future research. 
By enabling retrieval-free fact-checking, it allows for practical speed-ups and reduced reliance on external databases, while naturally complementing retrieval-based pipelines to further improve factual performance. 
Moreover, it provides a foundation for studying and strengthening LLMs’ intrinsic hallucination detection capabilities. 
Finally, this setting supports the development of \textit{factuality-oriented reward models}~\citep{stiennon2020learning,christiano2017deep}, which can be used as training signals in reinforcement learning or integrated directly into the generation process.

The contributions of the paper are as follows:
\begin{enumerate}[leftmargin=8pt, itemsep=0.3pt, topsep=0.2pt]

\item We introduce the fact-checking without retrieval setting, where the factuality of \emph{atomic claims} is assessed without external retrieval, using only an LLM’s internal knowledge.
    
\item We propose a large-scale evaluation framework spanning 9 datasets, designed to systematically assess robustness across diverse generalization dimensions.

\item We evaluate 18 methods with 3 models and introduce \textbf{INTRA}, an internal-based claim verifier that achieves SoTA average performance and demonstrates strong robustness. Additionally, we release data suite to support future research.\footnote{\url{https://huggingface.co/collections/s-nlp/intra}}

\end{enumerate}

\section{Related Work}
Traditional approaches to fact-checking and hallucination detection rely mainly on RAG. These systems verify the output of models by checking them against external knowledge sources~\citep{min-etal-2023-factscore,wei2024measuring,aushev-etal-2025-ragulator,rykov-etal-2025-models}. FActScore~\citep{min-etal-2023-factscore} breaks down the generated text into atomic facts and then calculates how many of them are actually supported by reliable sources such as Wikipedia. While this approach works quite well, RAG-based methods have several drawbacks: they require significant computational resources, their performance strongly depends on retrieval quality, and they are limited by the coverage of external knowledge bases. These limitations prevent them from fully utilizing the model's own parametric knowledge.

To address these limitations, another promising direction has emerged that uses LLM internal representations without needing any external retrieval. SAPLMA~\citep{azaria2023internal} showed that simple linear classifiers trained on hidden layer activations can distinguish true statements from false ones with 60-80\% accuracy. ~\citet{DBLP:conf/iclr/OrgadTGRSKB25} found that information about truthfulness tends to concentrate in specific tokens and layers. Interestingly, models might encode correct answers internally even when they generate incorrect ones. However, these methods often struggle when applied to different domains than they were trained on.

A related line of research focuses on uncertainty quantification methods~\citep{DBLP:conf/iclr/KuhnGF23,DBLP:conf/emnlp/FadeevaVTVPFVGP23,10.1162/tacl_a_00737,belikova-etal-2024-jellybell}, which try to analyze how confident a model is. Traditional approaches often mix different types of uncertainty that do not really relate to factuality~\citep{DBLP:journals/nature/FarquharKKG24}. Claim Conditioned Probability (CCP)~\citep{fadeeva2024fact} addresses this by separating uncertainty about claim values from uncertainty about surface forms, and shows particularly good performance across different languages. Among recent supervised methods, UHead~\citep{shelmanov2025headpredictheadquestion} uses trainable attention-based heads, while other approaches rely on token-level Mahalanobis distance~\citep{vazhentsev2025token}. 

RAUQ~\citep{rauq} takes a different approach by identifying attention heads that consistently drop their activation when the model generates incorrect information.
Additionally, several contrastive and self-correction approaches have been proposed. Contrastive methods such as CCS~\citep{burns2022discovering} use contrastive learning objectives to learn representations of truthfulness, while DoLa~\citep{DBLP:conf/iclr/ChuangXLKGH24} improves truthful generation by contrasting different model layers during inference. Self-correction approaches try to iteratively refine outputs, though detailed analysis by ~\citet{DBLP:journals/tacl/KamoiZZHZ24} shows that self-correction without external feedback typically does not work well. In this context, our factuality detector can be understood as a specialized type of reward model~\citep{stiennon2020learning,christiano2017deep} that focuses on evaluating truthfulness.

\section{Approach}

\subsection{Task Description}

We define the task of \textbf{fact-checking without retrieval} below. Let a claim be a declarative statement represented by a sequence of tokens $\mathbf{y} = y_1, \dots, y_n$, where the source of the claim is not restricted and may include human-written, machine-generated, or mixed-origin text. The objective is to produce a truthfulness score $s \in [0, 1]$ that estimates the probability of the claim being factually correct, i.e., $s \approx P(\text{Verified} \mid \mathbf{y})$. Consequently, the truthfulness score should be defined solely as a function of the claim text, without conditioning on the surrounding context, or the original input and the full model output.

The verification function $f$ must operate without access to \textit{any} external knowledge, including web search results, retrieved documents from a vector database, or any other form of external evidence. This restriction is motivated by the fact that retrieval-based pipelines introduce substantial computational overhead and additional sources of error. In practice, document retrieval typically requires multiple stages, including query formulation, nearest-neighbor search over large-scale indices, and re-ranking, which can add hundreds of milliseconds to several seconds per claim, depending on the corpus size and retrieval strategy \cite{DBLP:journals/corr/abs-2410-12837}. Moreover, retrieval quality is sensitive to query formulation, index coverage, and data freshness, making reliable and scalable deployment challenging \cite{percin-etal-2025-investigating}. Therefore, the assessment of truthfulness must be based solely on the parametric knowledge encoded within the model $M$ and the internal representations it generates when processing the claim $\mathbf{y}$.

Let us consider the following claim:
\textit{``The Eiffel Tower is located in Paris.''}
A retrieval-free verifier must evaluate this statement without querying an external database for the Eiffel Tower's location. Instead, it must infer the claim's veracity by analyzing the model's internal signals, such as hidden state activations, attention patterns, or output probabilities that arise when processing the text. The output would be a single score (e.g., $s=0.98$) indicating a high likelihood of the claim being true.

\subsection{Existing Methods}
\label{sec:methods}


Although initially motivated by detecting errors in LLM outputs, hallucination detection methods are also applicable to verifying claims from any source. Existing approaches to hallucination detection --- and by extension fact-checking without retrieval --- can be broadly categorized into \textit{supervised} and \textit{unsupervised} approaches, which leverage either output probabilities or internal signals from the model. Below, we provide an overview of representative baselines from each category.

\noindent\textbf{Unsupervised methods.} 
Uncertainty quantification is a widely used signal for hallucination detection and fact-checking \cite{shelmanov-etal-2025-uncertainty}, based on the assumption that LLMs are less confident when producing incorrect information. In this work, we focus on probability-- and internal–based methods~\citep{10.1162/tacl_a_00737}. Sampling-based approaches are unsuitable for claim verification without access to both the input prompt (that produced the claim) and the full generation. 

We consider several straightforward uncertainty measures: \textbf{Sequence Probability (SP)} computes the probability of a generated sequence; \textbf{Perplexity (PPL)}, estimates the average inverse log-likelihood of tokens; and \textbf{Mean Token Entropy (MTE)} averages the predictive entropy across tokens~\citep{fomicheva2020unsupervised}. Among state-of-the-art unsupervised measures, we include \textbf{Focus}~\citep{zhang-etal-2023-enhancing-uncertainty} propagates uncertainty from previous tokens via attention weights; \textbf{Claim-Conditioned Probability (CCP)}~\citep{fadeeva2024fact} conditions on the type and meaning of the claim to ignore surface-form and ``what to talk about'' uncertainty; \textbf{Recurrent Attention-based Uncertainty Quantification (RAUQ)}~\citep{rauq} identifies uncertainty-aware attention heads and combines their signals with perplexity; \textbf{Attention Score}~\citep{NEURIPS2024_LLM} computes the sum of eigenvalues of attention matrices. \textbf{Verbalized score (Verb)} evaluates factuality using the generated verbalized score from the language model. We use 10-shot examples for each claim and a Chain-of-Thought instruction. We additionally evaluate a retrieval-augmented variant (\textbf{Verb + RAG}), where for each claim, we retrieve the top-5 search snippets using the Google Serper API\footnote{\url{http://serper.dev}} to serve as external evidence.


\noindent\textbf{Supervised methods.} 
A further category of methods involves training lightweight classifiers on LLM representations to predict factuality. Several representative approaches fall into this category. \textbf{SAPLMA} trains a linear probe on hidden states collected from the 16th layer model, identified as the most informative~\citep{azaria2023internal}. 
\textbf{Contrast-Consistent Search (CCS)} employs contrastive training with a margin loss over last-layer embeddings, where negatives are hallucinations and positives are true statements. We adapt this method using a relaxed loss condition~\citep{burns2022discovering}. 
\textbf{Mass Mean Probe (MM)} is a linear probing technique that learns a projection defined by the difference of class means in hidden state space~\citep{marks2023geometry}.
\textbf{MIND} improves upon previous probing methods by optimizing both the selection of embeddings and the training configuration of the linear model~\citep{su2024unsupervisedrealtimehallucinationdetection}.
\textbf{Sheeps}\footnote{Please note that ``Sheeps'' is our own convenient adaptation of the method's name, as~\citet{ch2023androids} do not assign an explicit name to their approach in the original paper.} is a probing-based approach that trains lightweight classifiers on hidden states using attention pooling to detect hallucinations in grounded generation tasks~\citep{ch2023androids}. 
\textbf{Supervised Average Token Relative Mahalanobis Distance (SATRMD)} adapts Mahalanobis distance by computing token-level distances across layers and averaging them over tokens to serve as features for a model~\citep{vazhentsev2025token}. 

Finally, several methods exploit attention weights. \textbf{Trainable Attention-based Dependency (TAD)} models conditional dependencies between generation steps, using attention features to estimate the gap between conditional and unconditional confidence, and propagates uncertainty from earlier tokens to detect long-sequence hallucinations efficiently~\citep{vazhentsev-etal-2025-unconditional}. \textbf{UHead} combines multiple unsupervised uncertainty estimators with a trainable Transformer-based head on top of the language model and is specifically designed for hallucination detection in long-form generations~\citep{shelmanov2025headpredictheadquestion}.

\subsection{Intrinsic Truthfulness Assessment}

Although these methods are effective in specific settings, they face several limitations. First, the performance of supervised methods often degrades in out-of-distribution scenarios~\citep{vazhentsev2025token}, which are crucial for real-world applications. Second, methods based on hidden states tend to focus on particular algorithmic features~\citep{zhang-etal-2023-enhancing-uncertainty,vazhentsev2025token}, layers~\citep{azaria2023internal,su2024unsupervisedrealtimehallucinationdetection}, or tokens~\citep{azaria2023internal}, which further restricts their generalizability. Third, some methods that rely on attention weights~\citep{vazhentsev-etal-2025-unconditional,chuang-etal-2024-lookback,shelmanov2025headpredictheadquestion} typically require access to both the input prompt and the full generation, which limits their applicability in scenarios such as claim verification, where no prior knowledge of the input is available. To address these limitations, we propose the \underline{\textbf{In}}trinsic \underline{\textbf{Tr}}uthfulness \underline{\textbf{A}}ssessment (\textbf{INTRA}) method, which integrates the most effective insights from prior approaches into a unified and generalizable fact-checking framework.

\noindent\textbf{Token and layer selection.} 
Early attempts at supervised hallucination detection relied either on sequence-level embeddings -- typically formed by averaging token-level hidden states~\citep{su2024unsupervisedrealtimehallucinationdetection} -- or on the hidden states of the first or last generated token~\citep{azaria2023internal}. More recent work has shown that this assumption does not always hold. Instead, it proposes leveraging all token-level hidden states, aggregated using token-level uncertainty scores~\citep{vazhentsev2025token} or supervised attention pooling~\citep{ch2023androids}. In our method, we focus on the strong generalization and relying on fitted token-level uncertainty scores may not be the optimal solution.


We compute a sequence-level embedding using a learnable parameter vector $\boldsymbol{\theta}$ for a given sequence $\mathbf{y} =y_1, y_2 \dots y_N$ of length $N$, with corresponding hidden states $\mathbf{h}_l(y_i)$ for the $i-$th token from the $l-$th layer, following the approach of~\citet{ch2023androids}.
\begin{equation}
    \mathbf{h}_l (\mathbf{y}) =\sum_{i=1}^N \alpha_{l, i} \mathbf{h}_l (y_i), \quad \alpha_{l, i}=\frac{\exp \left(\boldsymbol{\theta}^{\top} \mathbf{h}_l (y_i)\right)}{\sum_{k=1}^N \exp \left(\boldsymbol{\theta}^{\top} \mathbf{h}_l (y_k)\right)}.
\end{equation}

Here, $\alpha_{l,i}$ represents the attention weight assigned to the hidden state of token $y_i$, normalized across the sequence via a softmax.

\noindent\textbf{Layer-wise truthfulness score.} To perform layer-wise claim verification, we apply a linear classifier with learnable weights $\mathbf{W}$ on top of the sequence-level embeddings from each layer:
\begin{equation}
    p_l(\text{Verified} \mid \mathbf{y})=\sigma\left(\mathbf{W}^{\top} \mathbf{h}_l (\mathbf{y})\right),
\end{equation}

where $\sigma(\cdot)$ is the sigmoid function, and $p_l(\text{Verified} \mid \mathbf{y})$ represents the probability that the sequence $\mathbf{y}$ is truthful according to layer $l$. We avoid complicating the training procedure or model architecture to ensure that the layer-wise scores do not overfit to specific patterns and remain broadly generalizable. The layer-wise models are trained using the standard cross-entropy loss.

\noindent\textbf{Aggregated truthfulness score.} Claim verification probabilities can be computed at various layers within a model. \citet{azaria2023internal,servedio-etal-2025-hidden} show that the optimal layer for this task can differ depending on the specific generation task. To effectively integrate information across layers, we follow the approach of~\citet{vazhentsev2025token} and train a regression model on top of the layer-wise probabilities. However, acknowledging that previous work has shown the first and last layers to be less effective, we use only the middle layers. We further argue that raw probabilities are not standardized across layers, which could degrade the performance of the regressor. Therefore, we apply quantile normalization~\citep{Amaratunga01122001} as $q(\cdot)$ to the probabilities before using them in the $\mathcal{L}_2$ regression:
\begin{equation}
\text{\textbf{INTRA}}(\mathbf{y}) = \sum_{l \in \mathcal{L}} \beta_l \cdot q\left(p_l(\text{Verified} \mid \mathbf{y})\right) + b,
\end{equation}
where $\beta_l$ and $b$ are the learnable weights and bias term, respectively, of the $\mathcal{L}_2$ regression model. We split the entire training dataset into two parts: the first is used to fit the parameters $\boldsymbol{\theta}$ and $\mathbf{W}$, while the second is used to fit $\beta_l, l \in \mathcal{L}$ and $b$. We use the layers from the first third to the second third of the model (e.g., layers 11 to 22 for Llama 3.1-8B-Instruct). We present an ablation study with the various ranges of layers in $\mathcal{L}$ in Table \ref{tab:layers_results}.








\section{Fact-Checking without Retrieval Benchmark}

\subsection{Datasets} 
To systematically study \textbf{fact-checking without retrieval} at the level of {atomic claims}, with a particular focus on generalization, we introduce a comprehensive evaluation protocol spanning heterogeneous sources and domains. Our framework validates performance on 9 diverse datasets, collectively probing multiple dimensions of the task: (i) long-tail knowledge, (ii) variation in claim sources between human-authored and model-generated claims, (iii) multilingual claims, (iv) claims extracted from long-form generations, and (v) cross-model claims.

\subsubsection{Long-Tail Knowledge}
A central aspect of our evaluation is assessing fact-checking performance across the spectrum of knowledge popularity. To this end, we introduce two datasets—\textbf{Atomic Claim PopQA (AC-PopQA)} and \textbf{Atomic Claim Wild Hallucinations (AC-WH)} - constructed from existing benchmarks. 
\textbf{AC-PopQA} is derived from {PopQA}~\citep{mallen2023trustlanguagemodelsinvestigating}, a QA dataset annotated with entity popularity based on Wikipedia page views, enabling controlled analysis of long-tail knowledge. 
\textbf{AC-WH} is built from {Wild Hallucinations}~\citep{zhao2024wildhallucinationsevaluatinglongformfactuality}, which targets long-tail facts and requires long-form generation, allowing us to study factuality under extended generations.

To construct atomic claims, we start from the original queries and generate answers using \textit{Llama 3.1-8B-Instruct}. For short-form generations, answer correctness is assessed with \textsc{InAccuracy}~\citep{moskvoretskii2025adaptive}. For long-form generations, we first decompose each response into atomic claims and then validate each claim with \textit{Llama 3.1-70B-Instruct} similarly to FActScore~\citep{min-etal-2023-factscore}.

\subsubsection{Human-Authored and Model-Generated Claims}
Our benchmark further distinguishes between two claim sources: \emph{human-authored} and \emph{model-generated} claims. 
Human-made claims are drawn from existing crowd-sourced benchmarks. \textbf{AVeriTeC}~\citep{schlichtkrull2023averitec} consists of real-world factual claims verified by human annotators, while \textbf{X-Fact}~\citep{gupta2021x} is a crowd-sourced dataset providing multilingual claims across 25 languages.

To study generated claims, we include both rule-based and LLM-generated statements. The \textbf{Cities}~\citep{marks2023geometry}, \textbf{Companies}~\citep{azaria2023internal}, and \textbf{CounterFact}~\citep{meng2022locating} datasets are constructed from Wikipedia using simple rule-based templates, yielding claims such as ``The city of Zürich is in Switzerland'' or ``Johann von Rist works as a poet.'' In contrast, datasets such as \textit{UHead}~\citep{shelmanov2025headpredictheadquestion}, \textit{Common Claims}~\citep{casper2023exploreestablishexploitred}, \textit{AC-PopQA}, and \textit{AC-WH} consist of claims generated by LLM.

\subsubsection{Multilinguality}
To extend our evaluation to a multilingual setting, we include all 25 languages from \textbf{X-Fact}~\citep{gupta2021x}, spanning a range of language families and typological properties. 
It covers \emph{Indo-European} languages (e.g., German, Spanish, Russian), \emph{Indo-Aryan} languages (e.g., Hindi, Bengali), \emph{Turkic} languages (e.g., Turkish, Azerbaijani), as well as languages from the \emph{Afro-Asiatic}, \emph{Dravidian}, \emph{Kartvelian}, and \emph{Austronesian} families. This diversity enables a systematic analysis of fact-checking robustness across languages with varying morphology, scripts, and resource availability.

\subsubsection{Long-Form Generation Claims}
We further evaluate LLM-generated claims extracted from \emph{long-form generations}. In this setting, the model is prompted to produce an extended passage, from which individual atomic claims are subsequently extracted. This distinction is important, as claims appearing later in a generation may follow a different distribution than earlier ones. To capture this effect, we include claims from long-form generations in two datasets: \textbf{AC-WH}, focusing on long-tail knowledge, and \textbf{UHead}, which targets popular knowledge.

\subsubsection{Cross-model Claims}
A desirable property of a fact-checker is independence from the LLM that generated the claim, ensuring robustness against model-specific artifacts or exploits. To evaluate this, our benchmark includes claims produced by multiple LLMs. Specifically, \textit{Common Claims} are generated by \textit{GPT-3-davinci-002}, \textit{UHead} by \textit{Mistral-7B-Instruct-v0.2}, and both \textbf{AC-WH} and \textbf{AC-PopQA} by \textit{Llama 3.1-8B-Instruct}. This setup enables a systematic assessment of cross-model generalization for retrieval-free fact-checking.

\noindent\textbf{Claim filtering.}
We apply an additional filtering procedure with \textit{Llama 3.3 72b Instruct} to ensure that each claim is high-quality and self-contained, i.e., it contains all information required for fact-checking without reliance on surrounding context. Specifically, each claim should describe a unique characteristic of a single entity. As shown in Table~\ref{tab:filtered_examples}, some data is accurate, but due to the use of pronouns, it is difficult to verify the accuracy of the information without additional context.  Datasets statistics and filtering details are reported in Table~\ref{tab:filtered_data}.

\subsection{Metrics} 
We use ROC-AUC and PR-AUC as the primary evaluation metrics. Both are threshold-independent and evaluate how well a method ranks hallucinated claims (label~1) above truthful ones (label~0). ROC-AUC measures overall separability, while PR-AUC focuses on the quality of hallucination predictions. In particular, PR-AUC reflects how often claims predicted as hallucinations are truly incorrect, which is important to avoid incorrectly flagging correct information as hallucinated. This is especially relevant for modern LLM case when hallucinations are relatively rare and over-predicting them would make a fact-checker impractical.

\subsection{Training Part}
To reserve the training part we divide AC-PopQA into train and test by stratifying on popularity, so that both sets maintain a similar distribution of popular and less popular entities. To avoid domain adaptation effects, we also ensure that entities and question templates from the training set do not appear in the test set. This set is later used for training all trainable methods. Other datasets in our benchmark do not include training set. 

\subsection{Models} 
We primarily evaluate claim verification methods on open-source language models, including \textit{Llama~3.1-8B Instruct}~\cite{llama3}, \textit{Ministral~8B Instruct 2410}~\cite{liu2026ministral3}, and \textit{Phi~4-mini Instruct}~\cite{DBLP:journals/corr/abs-2503-01743}, for both probability-based and embedding-based approaches. We also report an upper-bound reference by fact-checking with \textit{GPT-4.1}~\cite{DBLP:journals/corr/abs-2303-08774}, while noting its large size (approx. $>$120B), high cost, and reliance on a closed-source system.


\begin{table}[t!]
\caption{Average performance of claim verification methods for considered models in the proposed retrieval-free setting, measured by PR-AUC$\uparrow$ and ROC-AUC$\uparrow$ across nine datasets. \textbf{Bold} values indicate the best-performing retrieval-free method for each model, the second best is \underline{underlined}. \textit{Avg} column reports the average score across all models, summarizing overall robustness.}
\label{tab:overall}
\centering
\resizebox{0.85\columnwidth}{!}{\Large\begin{tabular}{l|ccc|c||ccc|c}
\toprule
 \multirow{2}{*}{\textbf{Method}} & \multicolumn{4}{c||}{\textbf{ROC-AUC}} & \multicolumn{4}{c}{\textbf{PR-AUC}} \\
 & \textbf{Llama 3.1} & \textbf{Ministral} & \textbf{Phi-4} & \textbf{Avg} & \textbf{Llama 3.1} & \textbf{Ministral} & \textbf{Phi-4} & \textbf{Avg} \\\midrule
 \multicolumn{9}{c}{\textit{Retrieval-Based Methods}} \\
\midrule

Verb+RAG & 77.7 & 78.2 & 76.5 & 77.5 & 70.1 & 69.6 & 67.1 & {68.9} \\
\midrule

 \multicolumn{9}{c}{\textit{Retrieval-Free Unsupervised Methods}} \\
\midrule
SP & 70.9 & 69.0 & 66.4 & 68.8 & 66.9 & 64.9 & 62.8 & 64.9 \\
PPL & 68.6 & 67.3 & 66.6 & 67.5 & 65.6 & 64.2 & 63.1 & 64.3 \\
MTE & 62.5 & 60.9 & 60.8 & 61.4 & 59.0 & 58.0 & 56.6 & 57.9 \\
Att. Score & 50.0 & 51.1 & 50.3 & 50.5 & 50.2 & 51.0 & 50.6 & 50.6 \\
RAUQ & 63.9 & 63.8 & 61.6 & 63.1 & 61.1 & 60.7 & 58.6 & 60.2 \\
CCP & 69.8 & 67.1 & 48.6 & 61.9 & 67.6 & 65.5 & 49.2 & 60.8 \\
Focus & 63.6 & 61.9 & 50.2 & 58.6 & 60.6 & 58.9 & 55.1 & 58.2 \\
Verb & 74.7 & \underline{72.2} & 69.1 & \underline{72.0} & 68.4 & 66.5 & 64.5 & 64.5 \\
\rowcolor{gray!10}

\textit{GPT-4.1} & - & - & - & \textit{84.0} & - & - & - & \textit{78.3} \\
\midrule
\multicolumn{9}{c}{\textit{Retrieval-Free Supervised Methods}} \\
\midrule
UHead & 63.1 & - & - & - & 60.2 & - & - & - \\
MM & 71.4 & 58.7 & \textbf{73.3} & 67.8 & 68.3 & 57.1 & \underline{68.7} & 64.7 \\
CCS & 71.1 & 53.0 & \underline{73.1} & 65.7 & 64.1 & 51.8 & \textbf{68.9} & 61.6 \\
TAD & 65.7 & 63.8 & 64.7 & 64.7 & 62.4 & 60.9 & 61.2 & 61.5 \\
SATRMD & 69.6 & 65.3 & 64.9 & 66.6 & 65.5 & 62.5 & 61.2 & 63.1 \\
MIND & 70.6 & 66.8 & {70.0} & 69.1 & 67.0 & 64.4 & {65.4} & 65.6 \\
Sheeps & \underline{75.0} & {71.6} & 69.4 & \underline{72.0} & \underline{70.5} & \underline{67.6} & {64.6} & \underline{67.6} \\
SAPLMA & 70.2 & 68.4 & 65.4 & 68.0 & 66.1 & 64.5 & 62.6 & 64.4 \\\midrule
INTRA & \textbf{77.7} & \textbf{{72.8}} & {69.5} & \textbf{{73.3}} & \textbf{73.1} & \textbf{{69.3}} & 64.6 & \textbf{69.0} \\
\bottomrule
\end{tabular}
}\end{table}

\section{Results}

Table~\ref{tab:overall} reports the average ROC-AUC and PR-AUC scores for claim verification methods in the proposed retrieval-free setting across nine datasets for all models.

Additionally, Tables~\ref{tab:main_rocauc} and~\ref{tab:pr_auc} report detailed ROC-AUC and PR-AUC results for Llama~3.1, Tables~\ref{tab:main_rocauc_ministral8b} and~\ref{tab:main_prauc_ministral8b} for Ministral, and Tables~\ref{tab:main_rocauc_phi4} and~\ref{tab:main_prauc_phi4} for Phi-4, for claim verification methods in the retrieval-free setting across nine datasets.

For the UHead baseline, we use the pre-trained open-source checkpoint\footnote{\url{https://hf.co/llm-uncertainty-head/uhead_claim_Llama-3.1-8B-Instruct}}, which is only provided for the Llama~3.1 model among the considered models~\citep{shelmanov2025headpredictheadquestion}.



In our experiments, PopQA is treated as the in-domain dataset, since the supervised methods are trained on its training split, while all other datasets are considered out-of-domain. The results highlight clear differences in robustness between the families of methods.  

\noindent\textbf{In-domain performance.} 
As expected, most supervised methods significantly outperform unsupervised retrieval-free and retrieval-based methods, including the GPT-4.1 baseline. The only exception is UHead, pre-trained but not fine-tuned on PopQA. The highest ROC-AUC is achieved by the proposed INTRA method, outperforming the second-best Sheeps method by 0.5\% on Llama~3.1.

\noindent\textbf{Unsupervised methods.} 
We also found that uncertainty-based approaches, with the exception of SP, generally underperform compared to other methods. The model’s raw confidence is not always well aligned with hallucinations on arbitrary inputs, a trend consistent with observations reported in standard uncertainty quantification studies for LLMs~\citep{10.1162/tacl_a_00737}.

Across evaluated baselines, we identify the Verbalized assessment as a promising direction with remarkably strong result, that stands out as the best-performing unsupervised approach, which contrasts with recent findings on uncertainty quantification~\citep{10.1162/tacl_a_00737}. However, it is significantly more compute-intensive than all other methods. Unlike the Verbalized approach, which requires generating multiple tokens to assess the target sequence, other methods complete the task with a single forward pass. Moreover, it suffers from a very high refusal rate on non-English inputs, with up to 58\% of cases resulting in refusals or similar behavior.


\noindent\textbf{Generalization performance.} 
Several baselines achieve strong results on specific datasets but fail to generalize across hallucination types. For example, MM performs well on its original benchmark but fails on long-form generations such as UHead or WH. Similarly, the UHead model leads in WH and its own dataset, yet lags behind in all other settings. This result illustrates the broader problem of low generalization capacity in hallucination detection~\citep{levinstein2024still}.

We also observe strong generalization from contrastive objectives, which yield competitive results in the specific settings and suggest that contrastive training can enhance robustness in this task, consistent with prior work in other domains~\citep{moskvoretskii2024mlem}.

Finally, our proposed detector, INTRA, addresses these issues by achieving the highest average performance and demonstrating consistent robustness across datasets. While not always the best-performing on individual benchmarks, it performs reliably in every setting. By combining token-level hidden states across layers, it captures rich internal information, which appear crucial for retrieval-free detection, consistent with prior findings~\citep{dombrowski2024information}. Overall, the proposed INTRA method outperforms the second-best retrieval-free Sheeps method by 2.7\% in ROC-AUC across datasets for Llama~3.1 and by 1.3\% across all models.


Overall, INTRA exhibits highly robust and consistent performance across both PR-AUC and ROC-AUC. While INTRA matches the Verb+RAG performance in ROC-AUC, it surpasses Verb+RAG by an average of 3\% in PR-AUC across datasets. Moreover, INTRA requires approximately 20x less computational time than the Verb+RAG method, as detailed in Appendix~\ref{sec:comp_time}.

\noindent\textbf{Saturation and contamination.} 
We also find that widely used rule-generated datasets are largely saturated, with trends diverging from those observed on other benchmarks. As these datasets have been repeatedly used in prior work~\citep{burns2022discovering,marks2023geometry}, we suspect contamination and overfitting at the method design stage. This is further supported by the MM method, which performs poorly across all datasets except the one on which it was originally introduced, indicating overfitting to leaked benchmarks.

\noindent\textbf{Downstream applications.} 
Main results suggest that training the full LLM, particularly with a mix of uncertainty-based objectives, can be highly effective. This is evidenced by the UHead method, which ranks first on long-form generation -- its original target domain. Moreover, its strong performance and generalization on WH, a dataset constructed with a different model and focused on long-tail entities, indicate substantial potential for scaling and broader applicability.

\begin{figure*}[t!]
    \centering
    \includegraphics[trim={0.32cm 0.25cm 0.25cm 0.3cm},clip,width=0.73\linewidth]{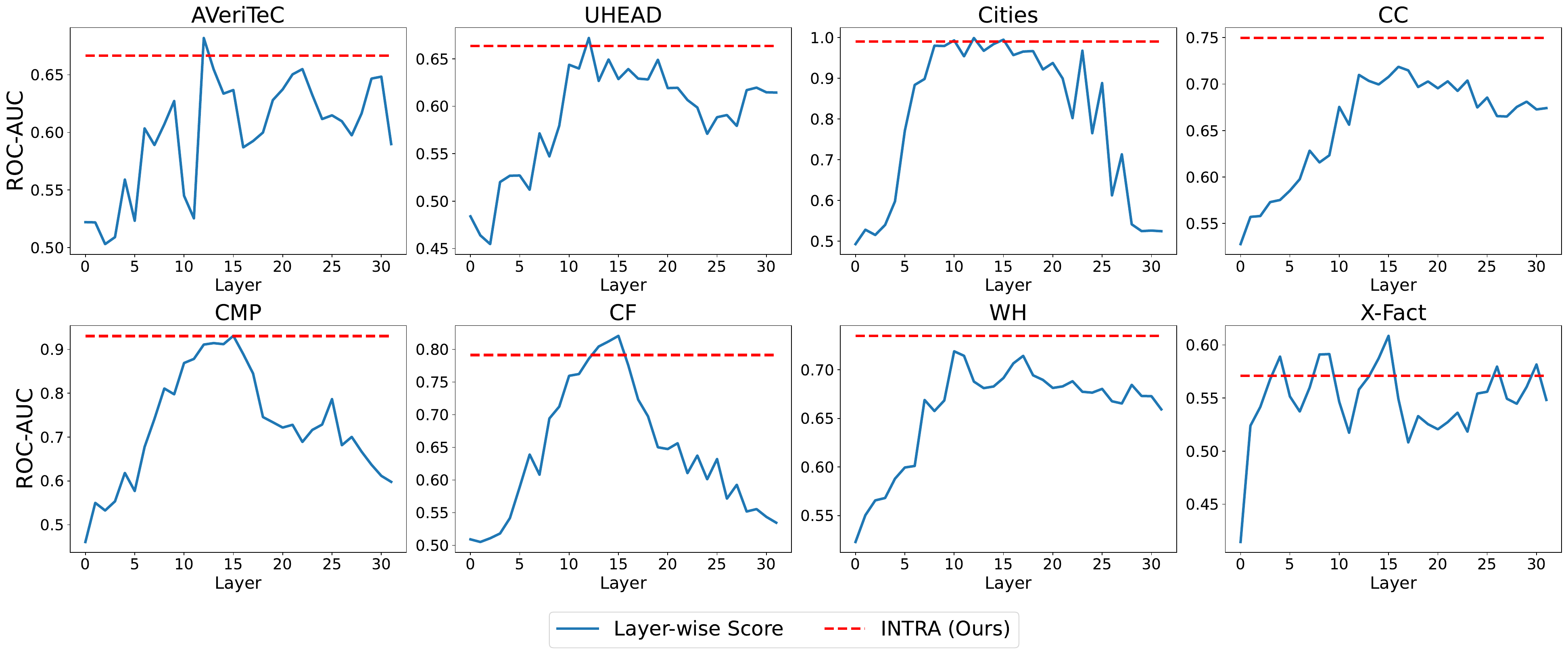}
    \caption{
    ROC-AUC$\uparrow$ performance of individual layers in the INTRA method.
    }
    \label{fig:layer_scores}
\end{figure*}

\section{Analysis}

\noindent\textbf{Performance of individual layers.} 
Figure~\ref{fig:layer_scores} presents the ROC-AUC performance of individual layers across various out-of-domain datasets compared to the full INTRA method. The results indicate that, across all tasks, intermediate layers generally yield better performance, consistent with prior work~\citep{azaria2023internal,vazhentsev2025token,servedio-etal-2025-hidden}. Although the most effective layer varies across datasets, the proposed INTRA method achieves performance equal to or exceeding that of the best individual layer, highlighting the advantage of integrating information across multiple layers.

\noindent\textbf{Layer selection.} 
Table~\ref{tab:layers_results} presents the results of the INTRA method trained on different subsets of the model’s hidden layers. We evaluate using all layers, the first layers (0–8), the last layers (24–32), and various ranges of middle layers. These results demonstrate that training only on the first or last layers is generally ineffective for claim verification tasks. In contrast, training on a small subset of intermediate layers typically yields optimal or near-optimal performance with minimal variation. Importantly, using a single middle layer performs significantly worse than training on a set of middle layers, which additionally underscores the importance of integrating information from multiple layers in a single score.

\begin{figure}[!t]
  \centering
  \begin{subfigure}{0.49\textwidth}
    \includegraphics[trim=0.5cm 0cm 0cm 0cm, width=\linewidth]{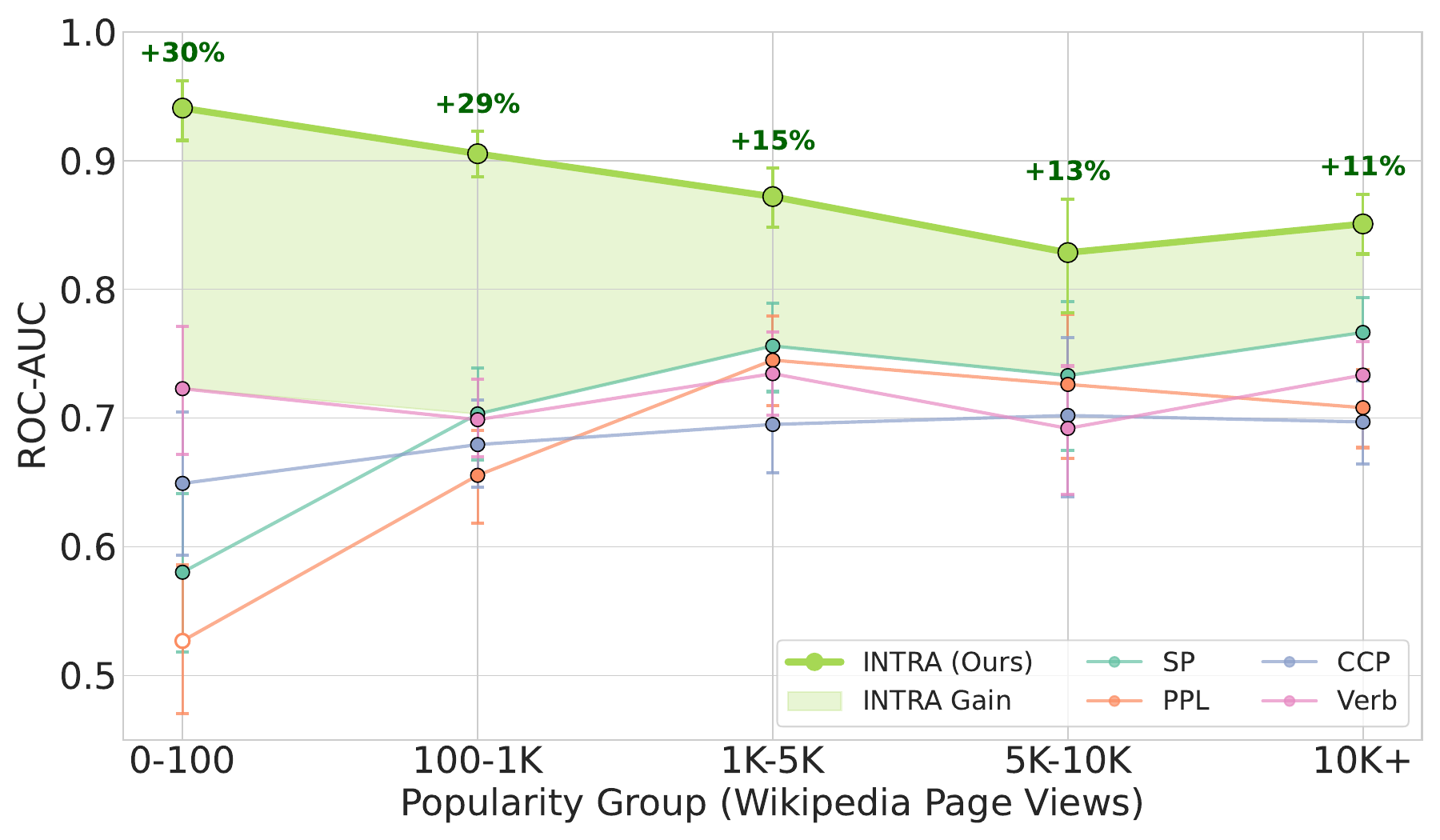}
    \caption{ROC-AUC on PopQA, split into five popularity groups. The green arrow shows the percent improvement of the top method (INTRA) over the runner-up.}
    \label{fig:popularity}
  \end{subfigure}
  \hfill
  \begin{subfigure}{0.49\textwidth}
    \includegraphics[width=\linewidth,height = \textheight, keepaspectratio, trim=0.3cm 0cm 1cm 0cm]{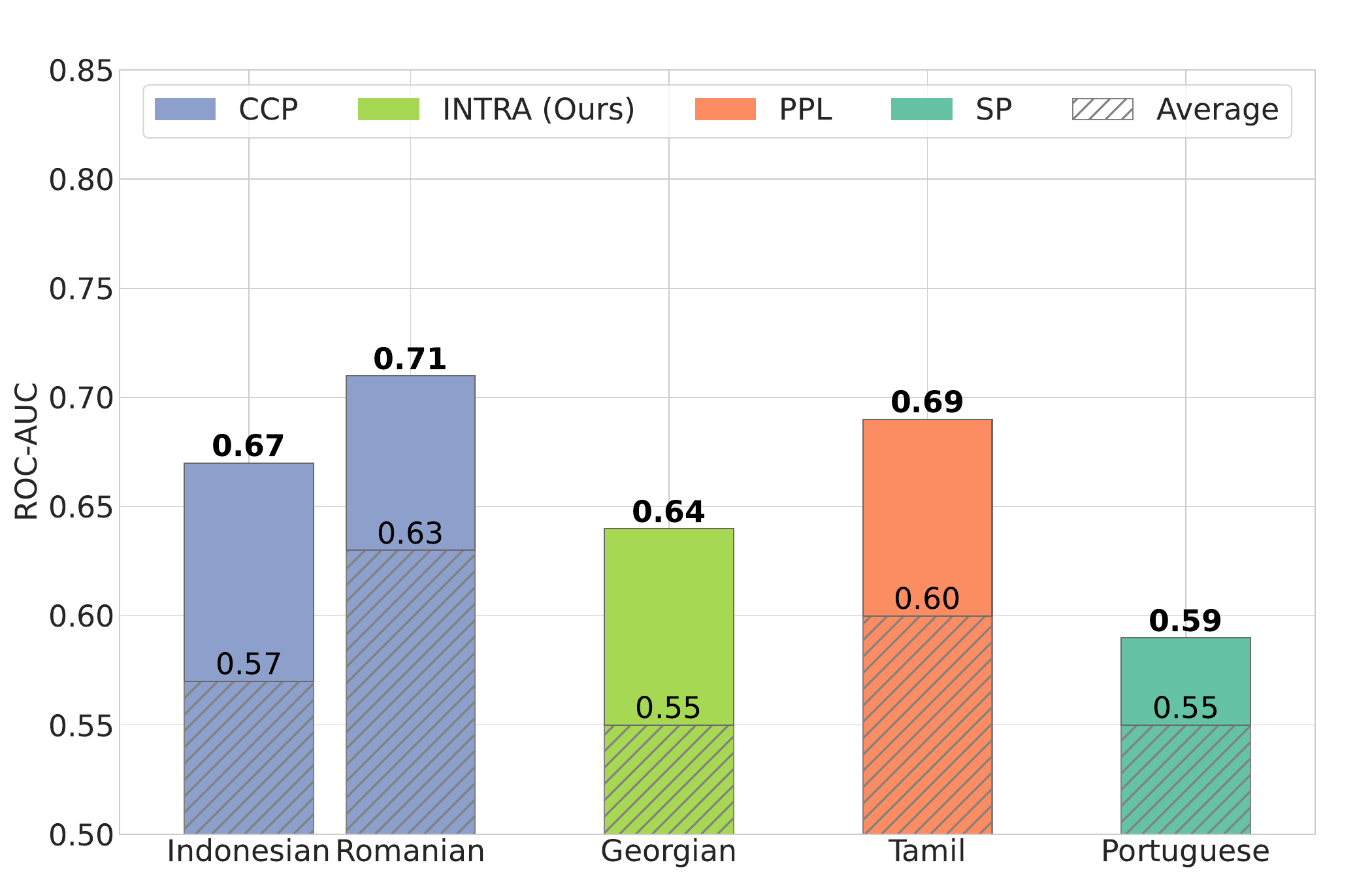}
    \caption{ROC-AUC on X-Fact by language. Each column shows the top method, with a hatched overlay indicating the mean of the other methods (SP, PPL, CCP, INTRA, Verb).}
    \label{fig:lang}
  \end{subfigure}
  \caption{ROC–AUC splited by popularity and language groups.}
  \label{fig:roc_auc_splits}
\end{figure}

\textbf{Long-tail performance.}
Figure~\ref{fig:popularity} presents results stratified by entity popularity. INTRA dominates the performance, indicating that supervised and internal signals enable robust hallucination detection even for rare entities. The verbalized detector is strongest for rare entities (0–100 group), but its advantage fades with increasing popularity. In contrast, SP improves as popularity grows, while both SP and PPL fail on the rarest entities, revealing their weakness on long-tail knowledge.

Confidence intervals were estimated via 2,000 bootstrap resamples. One-sided hypothesis tests indicate that INTRA achieves statistically significant gains across all popularity groups at the 95\% confidence level, with multiple comparisons controlled using the Benjamini–Hochberg false discovery rate (FDR) procedure.


\noindent\textbf{Language analysis.}
Due to the limited number of observations for many individual languages, we restrict the analysis to the subset with sufficient support to yield stable ROC-AUC estimates — Indonesian (id), Romanian (ro), Georgian (ka), Tamil (ta), and Portuguese (pt). Results for the remaining languages are not statistically reliable, so we do not interpret them further.

As shown in Figure~\ref{fig:lang}, the top-performing method varies by language. CCP achieves the highest ROC-AUC for Indonesian (0.67 vs. an across-method average of 0.57) and Romanian (0.71 vs. 0.63), indicating strong performance in these Latin-script settings. INTRA performs best for Georgian (0.64 vs. 0.55), suggesting robustness in a non-Latin, lower-resource setting. PPL reaches the highest ROC-AUC for Tamil (0.69 vs. 0.60), consistent with strong performance in non-Latin, higher-resource conditions, while SP is best for Portuguese (0.59 vs. 0.55). Across these five languages, the best method consistently outperforms the across-method average by roughly 0.04–0.10 ROC-AUC, indicating a meaningful advantage even when the winning approach differs by language.

Statistical significance ($p<0.10$) is assessed using one-sided bootstrap tests comparing the ROC-AUC for the top-performing method to the mean of ROC-AUC for other methods for that language.

\section{Conclusion}

We introduced the task of \textbf{fact-checking without retrieval}, systematically comparing 18 methods across nine datasets and proposing the INTRA approach, which achieves state-of-the-art results and strong generalization. Our experiments show that LLMs encode rich factuality signals in their internal representations that can be harnessed without retrieval, enabling lightweight and scalable claim verification. Beyond benchmarking, this paradigm opens avenues for integrating truthfulness signals directly into the generation process, serving as reward models for alignment or as monitoring modules in real-world deployments.

Our analyses highlight clear priorities for future work. \textbf{First}, \textit{middle layers} emerge as especially informative, suggesting that training should explicitly focus on these representations. \textbf{Second}, cross-dataset results reveal \textit{clear winners}: some methods consistently outperform in certain languages and on claims involving rare entities, showing that detector choice matters and that targeted selection can yield substantial gains. \textbf{Finally}, incorporating \textit{claim position and generation lengths} as structural features of long generations offers another promising direction for improving detection.

\bibliography{iclr2026_conference}
\bibliographystyle{iclr2026_conference}

\appendix
\clearpage
\section{Data Construction}
\label{sec:construct_data}

We study the generalization of hallucination detection across heterogeneous sources and domains. To ensure robustness and provide a comprehensive evaluation we propose the use of several datasets as a base for atomic claims:
\begin{itemize}[leftmargin=*]
    \item \textbf{PopQA}~\citep{mallen2023trustlanguagemodelsinvestigating}: This dataset contains simple rule-based questions annotated with a measure of popularity. It was designed to balance popular and less popular entities and to ensure robustness to social and cultural variation. We use it to evaluate model robustness to long-tail knowledge. For this purpose, we split the dataset into training and test sets by stratifying on popularity, so that both sets maintain a similar distribution of popular and less popular entities. To avoid domain adaptation effects, we also ensure that entities and question templates from the training set do not appear in the test set.  

    To construct atomic claims, we take the original questions, generate answers with \texttt{Llama-3.1-8B-Instruct}, and assess correctness using \textsc{InAccuracy}~\citep{moskvoretskii2025adaptive}. Each question–answer pair is then converted into an atomic claim with \texttt{Llama-3.1-70B-Instruct}, the hallucination label is defined by answers InAccuracy.

    The system prompt was ``You convert question-answer pairs into factual claims.'' and the few-shot prompt for a user prompt was as follows:

    \begin{tcolorbox}[colback=gray!5!white, colframe=gray!75!black, title= Converting QA pairs to a Claim Instruction]
    \label{popqa_prompt}
     
    Convert the question and answer into a factual, declarative English sentence.

    Examples:
    \vspace{1.1ex}
    
    Q: What sport does Kwak Hee-ju play?
    
    A: Table Tennis
    → Kwak Hee-ju plays table tennis.

     \vspace{1.1ex}
     
    Q: What is the religion of James VI and I?
    
    A: Protestantism
    → James VI and I confess protestantism.

     \vspace{1.1ex}
     
    Q: Who is the father of Sybilla of Normandy?
    
    A: Rollo of Normandy
    → The father of Sybilla of Normandy is Rollo of Normandy.
    
    \vspace{1.1ex}
     
    As an answer return just the resulting claim.
    
     \end{tcolorbox}
    
    \item \textbf{AVeriTeC}~\citep{schlichtkrull2023averitec}: A human-made fact-checking benchmark with a rigorous verification protocol. We adopt the validation split and use only the claims (without evidence), retaining only refuted or supported instances to reduce bias.

    \item \textbf{UHead}~\citep{shelmanov2025headpredictheadquestion}: An original dataset of questions about popular Wikipedia entities. Atomic claims are extracted with Gemma from long-form generations of a Mistral model, following the CCP setup~\citep{fadeeva2024fact}. This dataset tests cross-model robustness, since claims are generated by a different model, and also challenges robustness to long-form generation.
    
    \item \textbf{Cities}~\citep{marks2023geometry}:  a dataset formed of statements from the template ``\texttt{The city of [city] is in [country]}'' using a list of world cities. For each city, authors generated one true statement and one false statement.
    
    \item \textbf{Companies (CMP)}~\citep{azaria2023internal}: A dataset describing various general information about popular companies: headquarters location, what they do, well-known representatives etc.
    
    \item \textbf{Common Claims (CC)}~\citep{casper2023exploreestablishexploitred}: Dataset consists of various statements generated by GPT-3-davinci-002, labeled by humans as being true, false, or neither. The authors left only true/false variables and filtered out some facts to balance the dataset.
    
    \item \textbf{CounterFact (CF)}: Counterfact was introduced by ~\citep{meng2022locating} and consists of factual recall statements. The dataset consists of sentences constructed using various patterns: \texttt{[X] is [Y] citizen. [X] is owned by [Y]. [X] is the capital of [Y]} etc. Authors adapt Counterfact by using statements which form complete sentences and, for each such statement, using both the true version and a false version given by one of Counterfact’s suggested false modifications. 
    
    \item \textbf{Wild Hallucinations (WH)}~\citep{zhao2024wildhallucinationsevaluatinglongformfactuality}: The original dataset contains 7,917 user queries assuming long-form generation, covering a diverse set of domains. These queries correspond to real information needs but are not found in Wikipedia, thus representing long-tail knowledge. We generate answers to these questions with \texttt{Llama-3.1-8B-Instruct}, split them into atomic claims, and validate each claim against the provided evidence using FactScore~\citep{min-etal-2023-factscore}. The resulting atomic claims, labeled for hallucination by FactScore, form a dataset aimed at testing robustness to long-form generation in long-tail knowledge.

    \item \textbf{X-Fact}~\citep{gupta2021x}: The dataset consists of short statements in 25 languages, each labeled for hallucination by expert fact-checkers. It provides a multilingual evaluation benchmark designed to assess both out-of-domain generalization and the capabilities of multilingual models.
    
\end{itemize}

\section{Technical details}
\label{sec:technical}

Since our base model was \texttt{meta-llama/Llama-3.1-8B-Instruct} an instruction-tuned model, we extracted all claim embeddings using the chat template, with the user prompt set to ‘Generate true statement’ and the assistant content being the claim text itself.

\section{Data Filtering}
\label{sec:filtering}

Due to the lack of context in the task, we had to filter out some claims that could be correct, but did not meet the criteria. Specifically, each claim should describe a unique characteristic of a single entity. As shown in Table~\ref{tab:filtered_examples}, some data is accurate, but due to the use of pronouns, it is difficult to verify the accuracy of the information without additional context. Incorrect facts that appear to be correct in form were not filtered out (for example: Moby's album "Go" was released in 1992, not 1991). For filtering out, we used \texttt{Llama 3.3 72b Instruct} with prompt~\ref{filter_prompt} and 10-shot few-shot.

\begin{table*}[h]
\caption{Examples of filtered data. Claims generated by the model in response to the query: \emph{``Tell me facts about Moby, American musician and songwriter.''}}
\label{tab:filtered_examples}
\centering
\resizebox{0.95\textwidth}{!}{%
\begin{tabularx}{\textwidth}{@{} l c X @{}}
\toprule
\textbf{Claim} & \textbf{Filtered} & \textbf{Verification Explanation} \\
\midrule
Moby was born Richard Melville Hall. 
  & No 
  & This fact explicitly mentions the stage name ``Moby'' and provides a unique personal detail, his birth name. \\

Moby is an American singer-songwriter. 
  & No 
  & This fact explicitly mentions the entity ``Moby'' and describes his profession, directly tying the fact to him. \\


He rose to international fame. 
  & Yes 
  & The statement is too vague and lacks a specific entity or individual to whom it refers. \\

The album \emph{Play} came out in 1999. 
  & Yes 
  & This fact is vague as it doesn't specify the artist of the album ``Play,'' which could refer to multiple albums with the same title. \\

\bottomrule
\end{tabularx}
}
\end{table*}


\begin{table}[h]
\centering
\small
\caption{Final validation datasets after filtering claims. Cities, Common Claims, CounterFacts and Companies were not filtered out, as the claims are fully consistent with the original task.}
\label{tab:filtered_data}
\resizebox{0.7\columnwidth}{!}{
\begin{tabular}{llll}
\toprule
\textbf{Dataset}                 & \textbf{\# Before Filter} & \textbf{\# After Filter}  & \textbf{\% Filtered} \\
\midrule
AC-PopQA                    & 6,974  & 5,494  & 21          \\
AC-WH & 46,605 & 36,427 & 21          \\
AVeriTeC                 & 424    & 349    & 17          \\
UHead                    & 2,057  & 921  & 55          \\
Cities                   & 1,496      & 1,496  & 0           \\
Companies          & 1,200      & 1,200  & 0           \\
Common Claims       & 4,450      & 4,450  & 0           \\
CounterFact         & 1,200      & 1,200  & 0           \\
X-Fact                   & 2,315  & 1,131  & 51         \\
\midrule
Total  & 66,721  & 52,668  & 21 \\
\bottomrule
\end{tabular}
}
\end{table}




\begin{tcolorbox}[
    colback=gray!5!white,
    colframe=gray!75!black,
    title=Fact Filter Instruction,
    fonttitle=\bfseries,
    sharp corners,
    before skip=10pt,
    after skip=10pt
]
\label{filter_prompt}

You are an expert in verifying atomic facts extracted from Wikipedia. Your task is to determine whether a given fact refers exclusively to the specified topic or entity and cannot reasonably apply to others without direct reference.

\medskip

\textbf{Rules:}
\begin{itemize}[leftmargin=1.5em]
    \item The fact must be \textbf{unique} or \textbf{directly tied} to the target entity.
    \item Generic statements that could apply to many similar entities (e.g., people, places, events) are not acceptable.
    \item The entity must be \textbf{explicitly stated} in the fact.
    \item Whether the fact is true, false, or debatable is \textbf{not relevant}; only specificity matters.
    \item If the fact clearly names or describes something unique to the topic, it is \textbf{GOOD}.
    \item If the fact is vague, general, or potentially applies to many, it is \textbf{BAD}.
\end{itemize}

\medskip

\textbf{Output format:}  
Return a brief explanation (1–2 sentences), followed by a semicolon and then either \texttt{GOOD} or \texttt{BAD}.  
\textit{Example:}  
\texttt{Mentions a specific architectural style found only in this city; GOOD}
\end{tcolorbox}


To generate verbalized classifications, all considered models use the same prompt (Table~\ref{table:prompt_class}) and identical generation hyperparameters: \texttt{temperature=0.3}, \texttt{max\_tokens=150}, and \texttt{few\_shots=10}. Since verbalized evaluations can be unstable across generations, we use a relatively large number of few-shots and a low temperature to improve consistency across models.

\begin{tcolorbox}[
  colback=gray!5!white,
  colframe=gray!75!black,
  title=Verbalized Claim Classification,
  label={table:prompt_class},
  sharp corners,
  boxrule=0.8pt,
  fonttitle=\bfseries,
]
You are an expert fact-checker with comprehensive knowledge across all domains. Your task is to determine whether claims are factually \textbf{TRUE} or \textbf{FALSE} based on established facts and evidence.

\medskip
\textbf{Rules:}
\begin{itemize}
  \item Evaluate the factual accuracy of each claim based on verifiable information.
  \item Consider the claim exactly as written.
  \item Small errors in dates, numbers, or facts render the claim \textbf{FALSE}.
  \item If a claim contains multiple parts, \textbf{all parts must be true} for the claim to be \textbf{TRUE}.
  \item If any part is incorrect, misleading, or inaccurate, classify the claim as \textbf{FALSE}.
  \item Provide a brief explanation (1–2 sentences) of your reasoning.
  \item Output format: \textit{explanation;} followed by either \texttt{"TRUE"} or \texttt{"FALSE"}.
\end{itemize}
\end{tcolorbox}

\clearpage
\section{Computational Time.}
\label{sec:comp_time}
Table~\ref{tab:runtime_per_instance} shows the average per-instance runtime for all methods on PopQA using Ministral-8B-Instruct-2410 on an A100 GPU with batch size 1. All evaluated methods except the verbalized baseline require only a single forward pass of the LLM per claim, whereas standard RAG-based fact-checking pipelines involve hundreds or thousands of forward passes due to retrieval and re-evaluation. Probability-based approaches are the fastest overall, with runtimes around 0.02 seconds per instance. Supervised methods introduce only a small overhead, typically not exceeding 0.04 seconds, and therefore remain effectively lightweight. In contrast, the verbalized method is substantially slower, requiring 0.25 seconds on average with higher variance, making it considerably less practical. CCP is the most computationally expensive due to additional NLI-based computations. Overall, INTRA maintains a low runtime of 0.06 seconds per instance, confirming that it remains lightweight while offering strong performance.

\begin{table}[h!]
\centering
\small
\caption{Runtime per instance (in milliseconds) for different hallucination detection methods. Values are reported as mean $\pm$ standard deviation, measured per input instance.}
\label{tab:runtime_per_instance}
\setlength{\tabcolsep}{6pt}
\begin{tabular}{l c}
\toprule
\textbf{Method} & \textbf{Runtime per instance (ms)} \\
\midrule
\multicolumn{2}{c}{\textit{Retrieval-Based Unsupervised Methods}} \\  
\midrule
Verb+RAG      & $950 \pm 200$ \\
\midrule
\multicolumn{2}{c}{\textit{Retrieval-Free Unsupervised Methods}} \\  
 \midrule
SP         & $30 \pm 1$ \\
PPL        & $30 \pm 1$ \\
MTE        & $84 \pm 40$ \\
Att. Score & $40 \pm 10$ \\
RAUQ       & $33 \pm 10$ \\
CCP        & $563 \pm 220$ \\
Verb      & $250 \pm 110$ \\
\midrule
\multicolumn{2}{c}{\textit{Retrieval-Free Supervised Methods}} \\  
 \midrule
TAD        & $45 \pm 10$ \\
SATRMD     & $31 \pm 5$ \\
MIND       & $31 \pm 10$ \\
Sheeps     & $67 \pm 40$ \\
SAPLMA     & $31 \pm 5$ \\
\midrule
INTRA      & $56 \pm 20$ \\
\bottomrule
\end{tabular}
\vspace{-0.3cm}
\end{table}

\clearpage
\section{Ablation Studies}

\subsection{Layer Selection}
\label{sec:layers_selection}

Table~\ref{tab:layers_results} presents the results of the INTRA method trained on different subsets of the model’s hidden layers for the Llama model.

\begin{table}[!h] 

\caption{\label{tab:layers_results} Impact of different layer subsets on the performance by ROC-AUC$\uparrow$ of the proposed INTRA method. The best method is in \textbf{bold}, the second best is \underline{underlined}.}

\resizebox{\columnwidth}{!}{\begin{tabular}{l|ccccccccc|c}
\toprule
\textbf{Layers} & \textbf{PopQA} & \textbf{AVeriTeC} & \textbf{UHEAD} & \textbf{Cities} & \textbf{CC} & \textbf{CMP} & \textbf{CF} & \textbf{WH} & \textbf{X-Fact} & \textbf{Avg} \\\midrule
All & 89.5 & 66.4 & 63.7 & 98.7 & 74.2 & 89.1 & 73.7 & 73.3 & 58.1 & 76.3 \\
0-8 & 84.9 & 59.5 & 53.0 & 83.9 & 62.1 & 65.4 & 59.7 & 64.1 & 55.2 & 65.3 \\
24-32 & 86.7 & 63.8 & 61.0 & 73.2 & 71.5 & 70.5 & 58.4 & 70.0 & 57.0 & 68.0 \\
2-30 & \underline{89.6} & 66.2 & 63.6 & 98.8 & 73.8 & 89.7 & 74.7 & 73.1 & 58.3 & 76.4 \\
8-24 & \textbf{89.6} & \textbf{66.8} & 65.3 & \textbf{99.3} & 74.4 & 92.0 & 77.4 & \underline{73.5} & 57.6 & 77.3 \\
11-22 & 89.3 & \underline{66.7} & \textbf{66.4} & 99.0 & \textbf{75.0} & 93.0 & 79.1 & \textbf{73.5} & 57.1 & \underline{77.7} \\
13-19 & 88.8 & 65.4 & \underline{65.6} & \underline{99.0} & \underline{74.9} & \underline{94.0} & \underline{81.9} & 73.1 & \underline{58.7} & \textbf{77.9} \\
15-17 & 87.5 & 63.0 & 64.8 & 98.7 & 73.8 & \textbf{94.1} & \textbf{82.8} & 72.2 & \textbf{59.8} & 77.4 \\
16 & 87.2 & 58.7 & 63.9 & 95.7 & 71.9 & 88.9 & 77.6 & 70.7 & 54.9 & 74.4 \\
\bottomrule
\end{tabular}
}\end{table}

\subsection{Claim Position Influence}
\label{sec:uhead_claim_position}

\begin{figure}[htbp!]
  \centering
  \begin{subfigure}{0.44\textwidth}
    \includegraphics[trim=0cm 0.5cm 2cm 0cm, width=\linewidth]{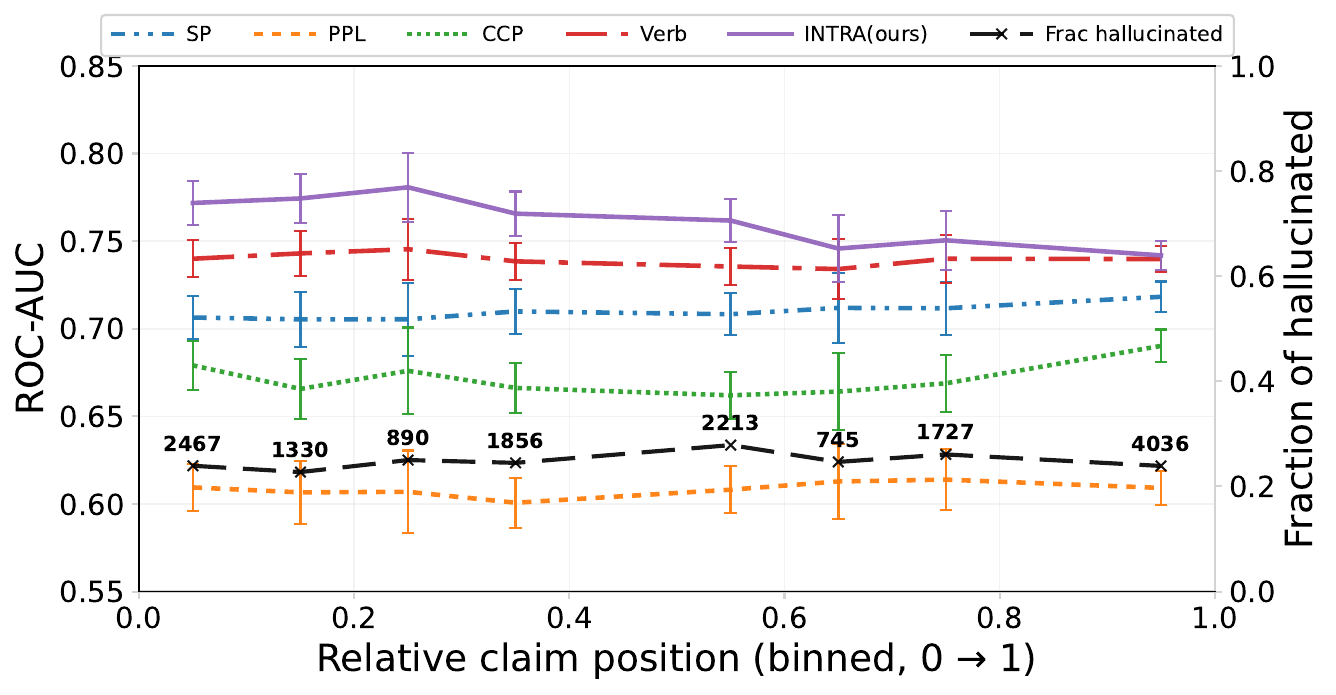}%
    \caption{Short-length claims.}
    \label{fig:position_short}
  \end{subfigure}
  \hfill
  \begin{subfigure}{0.47\textwidth}
    \includegraphics[width=\linewidth,height=\textheight, keepaspectratio, trim=1cm 0.45cm 0cm 1cm]{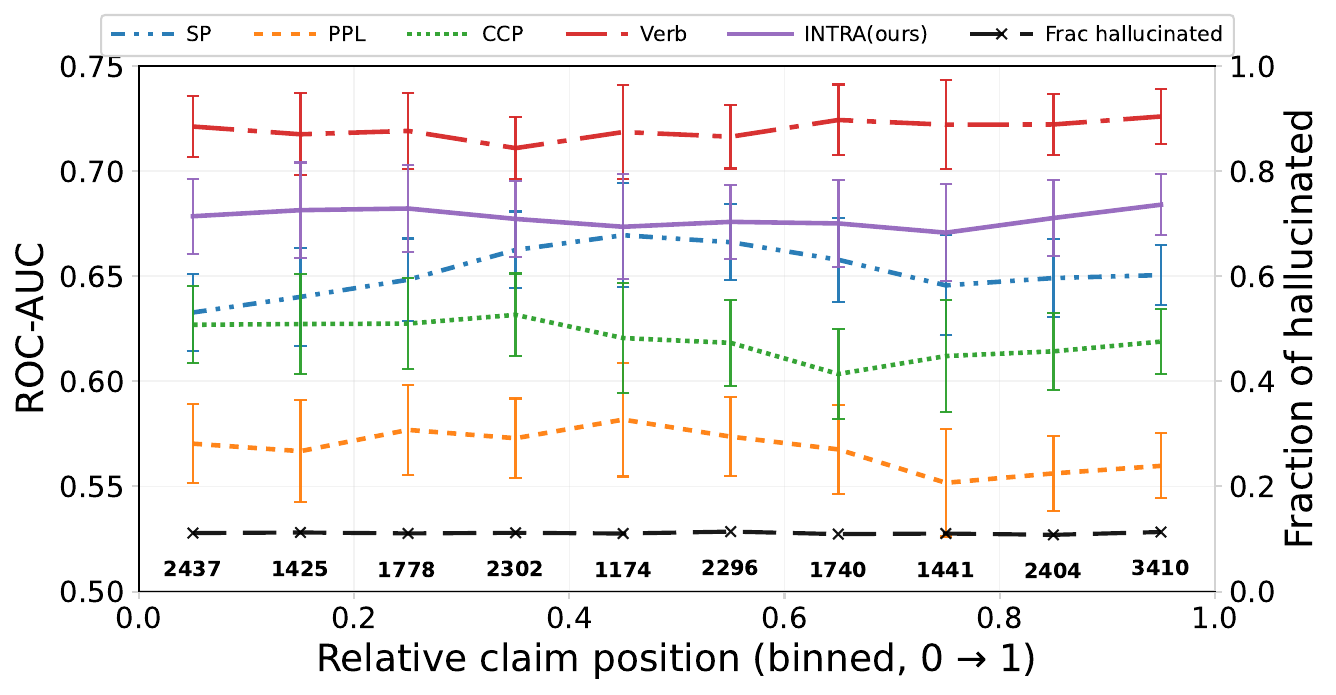}
    \caption{Medium-length claims.}
    \label{fig:position_medium}
  \end{subfigure}
  \caption{ROC–AUC on WH by claim position for short and medium-length claims. The dashed line shows the fraction of hallucinated claims at each position; the numbers above it indicate the number of claims at that position.}
  \label{fig:position}
\end{figure}

%
While standard evaluation metrics assume claim independence, datasets such as WH contain multiple claims generated within a single LLM output. To account for this structure, we compute ROC-AUC as a function of claim position and average results across generations, as shown in Figure \ref{fig:position}. Prior work has shown that hallucination rates can vary with claim position, often increasing toward later claims as errors accumulate~\citep{DBLP:conf/naacl/BelemPIMBH25,DBLP:conf/naacl/SpataruHVC24,DBLP:conf/acl/WangS20}.

Because generation lengths vary widely — from very short outputs with fewer than six claims to long generations containing dozens of claims — we partition generations into short (
$\le$ 6 claims), medium (7–12 claims), and long ($\ge$ 12 claims). To make positions comparable across different lengths, claim indices are normalized within each group and binned to relative positions in the range $[0, 1]$.


Figures~\ref{fig:position_short} show that INTRA consistently achieves higher ROC-AUC than the Verbalized baseline across most claim positions for short generations, with statistically significant improvements at early and late positions (approximately relative positions 0.15–0.35 and 0.55–0.95 after correction). In contrast, for medium-length generations (Figure ~\ref{fig:position_medium}), the Verbalized setting performs best and is statistically significantly better than INTRA at several early and late positions, while INTRA remains competitive at mid-claim positions. Notably, we do not observe a systematic increase in hallucination rates toward the end of generations, nor a degradation in detector performance at later claim positions.

Statistical significance ($p<0.05$) is assessed using one-sided bootstrap tests with FDR correction comparing the top-performing method to the second-best method at each claim position, long-length claims subset was not enough for statistical significance at any claim position.

\clearpage
\section{Detailed Experimental Results}
\label{sec:detailed_results}

\begin{table*}[h!]
\centering
\small
\caption{Performance of claim verification methods for \textbf{Llama-3.1-8B} in the proposed evidence-free setting, measured by ROC-AUC$\uparrow$ across nine datasets. \textbf{Bold} values indicate the best-performing method for each dataset, the second best is \underline{underlined}. \textit{Avg} column reports the average score across all datasets, summarizing overall robustness. 
}
\label{tab:main_rocauc}
\setlength{\tabcolsep}{3pt}
\begin{tabular}{l|cccccccccc}
\toprule
\multirow{2}{*}{\textbf{Method}}
& \textit{\scriptsize long-tail}
& \textit{\scriptsize human-made}
& \textit{\scriptsize cross-model}
& \multicolumn{4}{c}{\textit{\scriptsize rule-generated}}
& \textit{\scriptsize long-form}
& \textit{\scriptsize multilingual}
& 
\multirow{2}{*}{\textbf{Avg}}
\\
\cmidrule(lr){5-8}
 & \textbf{PopQA} & \textbf{AVeriTeC} & \textbf{UHead} & \textbf{Cities} & \textbf{CC} & \textbf{CMP} & \textbf{CF} & \textbf{WH} & \textbf{X-Fact} &  \\

 \midrule
\multicolumn{11}{c}{\textit{Retrieval-Based Methods}} \\  
 \midrule

Verb + RAG & 72.8 & {74.4} & {72.7} & 98.4 & 73.3 & 89.0 & 74.3 & {71.0} & {73.3} & {77.7} \\

\midrule
\multicolumn{11}{c}{\textit{Retrieval-Free Unsupervised Methods}} \\  
 \midrule
SP & 78.2 & \underline{67.4} & 69.9 & 95.8 & 65.0 & 74.9 & 70.1 & 70.0 & 53.5 & 71.6\\
PPL & 73.3 & 57.4 & 62.4 & 94.2 & 72.5 & 75.0 & 65.9 & 59.3 & 57.6 & 68.6\\
MTE & 65.9 & 66.6 & 60.7 & 56.8 & 69.0 & 64.3 & 52.5 & 66.1 & \textbf{61.1} & 62.6 \\
Att. Score & 49.7 & 59.4 & 58.6 & 41.5 & 42.8 & 47.8 & 49.1 & 58.6 & 42.7 & 50.0\\
RAUQ & 65.6 & 67.3 & 61.2 & 57.0 & 68.4 & 63.6 & 52.4 & 67.3 & \underline{60.9} & 62.6\\
CCP & 72.5 & 65.1 & 69.1 & 89.6 & 65.9 & 79.3 & 66.5 & 69.1 & 51.3 & 69.8 \\
Focus & 67.3 & 59.4 & 58.0 & 74.2 & 63.1 & 66.5 & 59.8 & 67.8 & 55.8 & 63.5 \\
Verb & 72.8 & 62.9 & \textbf{72.8} & 98.6 & \underline{74.3} & \underline{89.5} & 75.4 & \underline{73.8} & 51.9 & 74.7 \\
\rowcolor{gray!10}
\textit{GPT-4.1} & \textit{84.7} & \textit{76.7} & \textit{87.0} & \textit{99.5} & 8\textit{0.3} & \textit{92.3} & \textit{90.8} & \textit{81.8} & \textit{62.8} & \textit{84.0} \\
\midrule
\multicolumn{11}{c}{\textit{Retrieval-Free Supervised Methods}} \\  
 \midrule
UHead & 65.7 & 52.1 & \underline{71.2} & 61.6 & 68.9 & 66.9 & 54.5 & \textbf{74.2} & 52.6 & 63.1 \\
MM & 79.5 & 63.0 & 57.4 & \textbf{99.9} & 67.9 & 75.1 & \textbf{82.1} & 56.3 & 51.6 & 71.4 \\
CCS  & 86.6 & 54.1  & 66.5 & 95.9 & 67.9 & 77.6 & 66.2 & 66.5 & 54.4 & 71.1 \\
ICR & 74.9 & 51.0 & 70.9 & 58.3 & 50.7 & 54.9 & 51.4 & 56.5 & {57.7} & 58.5 \\
TAD & 84.4 & 56.5 & 60.1 & 73.1 & 65.3 & 69.6 & 53.6 & 70.6 & {57.8} & 65.7\\
SATRMD & 81.3 & \textbf{68.3} & 62.9 & 86.6 & 69.9 & 78.8 & 60.6 & 64.3 & 53.6 & 69.6\\
MIND & 88.7 & {66.8} & 64.5 & 91.3 & 70.0 & 71.2 & 66.2 & 65.9 & 50.5 & 70.6 \\
Sheeps & \underline{88.8} & 63.6 & 64.8 & 98.1 & 73.4 & 84.5 & 72.7 & 72.1 & 57.0 & \underline{75.0} \\
SAPLMA & 88.6 & 62.9 & 63.5 & 81.8 & 70.9 & 80.6 & 60.6 & 67.6 & 55.0 & 70.2\\
\midrule
\textbf{INTRA} & \textbf{89.3} & 66.7 & 66.4 & \underline{99.0} & \textbf{75.0} & \textbf{93.0} & \underline{79.1} & 73.5 & 57.1 & \textbf{77.7} \\
\bottomrule
\end{tabular}
\vspace{-0.5cm}
\end{table*}

\begin{table*}[h!]
\centering
\caption{
Performance of claim verification methods for \textbf{Llama-3.1-8B}  in the proposed evidence-free setting, measured by PR-AUC$\uparrow$ (multiplied by 100) across nine datasets. \textbf{Bold} values indicate the best-performing method for each dataset, the second best is \underline{underlined}. \textit{Avg} column reports the average score across all datasets, summarizing overall robustness.}
\label{tab:pr_auc}

\resizebox{\textwidth}{!}{
\begin{tabular}{lcccccccccc}
\toprule
& \textit{\scriptsize long-tail}
& \textit{\scriptsize human-made}
& \textit{\scriptsize cross-modal}
& \multicolumn{4}{c}{\textit{\scriptsize rule-generated}}
& \textit{\scriptsize long-form}
& \textit{\scriptsize multilingual}
& \\
\cmidrule(lr){5-8}
Method & PopQA & AVeriTeC & UHead & Cities & CC & CMP & CF & WH & X-Fact & Avg \\
\midrule
\multicolumn{11}{c}{\textit{Retrieval-Based Methods}} \\  
 \midrule
 Verb + RAG & 86.0 & {82.5} & {45.5} & 97.1 & 67.0 & {83.6} & {66.2} & 36.0 & 67.0 & 70.1 \\

\midrule
\multicolumn{11}{c}{\textit{Retrieval-Free Unsupervised Methods}} \\  
 \midrule

SP          & 89.4 & \textbf{79.6} & 40.4 & 94.4 & 64.1 & 71.5 & 66.5 & 32.0 & 64.4 & 66.9 \\
PPL          & 88.3 & 74.3 & 31.5 & 94.4 & 71.7 & 77.3 & 62.4 & 21.3 & 69.4 & 65.6 \\
MTE          & 83.7 & 78.5 & 30.6 & 55.2 & 66.6 & 61.4 & 51.6 & 29.4 & \textbf{73.9} & 59.0 \\
Att. Score   & 75.1 & 75.0 & 30.6 & 45.6 & 44.6 & 48.0 & 49.4 & 22.9 & 60.9 & 50.2 \\
RAUQ         & 83.6 & 79.1 & 31.0 & 55.2 & 66.2 & 61.3 & 51.6 & 30.6 & \underline{73.7} & 59.1 \\
CCP          & 87.7 & 78.8 & 44.8 & 90.5 & 65.7 & 79.0 & 64.7 & 32.4 & 65.0 & 67.6 \\
Focus        & 83.5 & 75.3 & 32.1 & 71.0 & 62.1 & 63.9 & 57.0 & 31.7 & 68.8 & 60.6 \\
Verb & 85.4 & 75.8 & 40.5 & 97.4 & 66.9 & \underline{84.3} & {67.2} & 32.0 & 66.0 & 68.4 \\
\rowcolor{gray!10}
\textit{GPT-4.1} & \textit{91.7}    & \textit{83.4} & \textit{62.8} & \textit{99.1}    & \textit{73.3}    & \textit{89.0}    & \textit{85.1}    & \textit{49.1}   & \textit{71.5} & \textit{78.3} \\
\midrule
\multicolumn{11}{c}{\textit{Retrieval-Free Supervised Methods}} \\
\midrule
UHead        & 84.1 & 70.7 & \textbf{62.6} & 58.3 & 65.2 & 65.0 & 53.0 & 36.5 & 64.0 & 60.2 \\
MM           & 92.7 & 78.6 & 31.7 & \textbf{99.9} & 68.3 & 76.9 & \textbf{80.3} & 21.1 & 64.8 & 68.3 \\
CCS          & 95.3 & 59.8 & 30.9 & 96.3 & 62.6 & 64.1 & 67.2 & \textbf{67.1} & 41.4 & 64.1 \\
ICR          & 89.5 & 71.6 & \underline{47.0} & 56.2 & 50.4 & 45.8 & 51.2 & 19.4 & 58.3 & 54.4 \\
TAD          & 94.0 & 75.4 & 31.2 & 72.2 & 63.1 & 66.4 & 52.7 & 37.8 & 68.6 & 62.4 \\
SATRMD       & 92.3 & 79.2 & 32.4 & 86.6 & 68.4 & 76.8 & 58.1 & 27.2 & 68.2 & 65.5 \\
MIND         & \underline{95.8} & \textbf{82.0} & 34.5 & 90.9 & 69.8 & 69.2 & 61.9 & 31.9 & 67.0 & 67.0 \\
Sheeps       & 94.9 & 76.8 & 35.1 & {97.8} & \underline{72.6} & 83.9 & 66.9 & 37.2 & 69.0 & \underline{70.5} \\
SAPLMA       & 96.0 & 73.9 & 35.9 & 80.9 & 69.2 & 81.8 & 57.4 & 32.7 & 67.5 & 66.1 \\
\midrule
\textbf{INTRA}        & \textbf{96.2} & 78.6 & 36.7 & \underline{99.3} & \textbf{73.5} & \textbf{92.5} & \underline{71.1} & \underline{41.8} & 68.6 & \textbf{73.1} \\

\bottomrule
\end{tabular}
}
\end{table*}

\begin{table*}[h!]
\centering
\caption{Performance of claim verification methods for \textbf{Ministral-8B}  in the proposed evidence-free setting, measured by ROC-AUC$\uparrow$ across nine datasets. \textbf{Bold} values indicate the best result per dataset, the second best is \underline{underlined}, and \textit{Avg} the mean score across all datasets.}
\label{tab:main_rocauc_ministral8b}
\resizebox{\textwidth}{!}{
\begin{tabular}{l|cccccccccc}
\toprule
\multirow{2}{*}{\textbf{Method}}
& \textit{\scriptsize long-tail}
& \textit{\scriptsize human-made}
& \textit{\scriptsize cross-model}
& \multicolumn{4}{c}{\textit{\scriptsize rule-generated}}
& \textit{\scriptsize long-form}
& \textit{\scriptsize multilingual}
& \multirow{2}{*}{\textbf{Avg}}
\\
\cmidrule(lr){5-8}
 & \textbf{PopQA} & \textbf{AVeriTeC} & \textbf{UHead} & \textbf{Cities} & \textbf{CC} & \textbf{CMP} & \textbf{CF} & \textbf{WH} & \textbf{X-Fact} &  \\
 \midrule
\multicolumn{11}{c}{\textit{Retrieval-Based Methods}} \\  
 \midrule
 Verb + RAG & 76.0 & 77.0 & 75.0 & 98.0 & 73.0 & 87.0 & 75.0 & 70.0 & 73.0 & 78.2 \\
\midrule
\multicolumn{11}{c}{\textit{Retrieval-Free Unsupervised Methods}} \\
\midrule
SP & 75.3 & \textbf{65.0} & \textbf{68.8} & 95.9 & 63.9 & 67.6 & 64.7 & \textbf{72.2} & 47.8 & 69.0 \\
PPL & 72.7 & 55.3 & 56.7 & 95.1 & \underline{71.4} & 70.3 & 62.3 & 61.0 & \textbf{61.3} & 67.3 \\
MTE & 66.4 & 62.3 & 59.1 & 52.8 & 69.7 & 56.7 & 51.0 & 69.1 & \textbf{61.3} & 60.9 \\
Att. Score & 50.4 & 59.6 & 59.8 & 46.5 & 43.5 & 48.7 & 49.5 & 59.2 & 42.9 & 51.1 \\
RAUQ & 68.2 & 55.9 & 56.1 & 88.9 & 68.6 & 57.7 & 56.3 & 60.9 & \underline{61.2} & 63.8 \\
CCP & 69.9 & 60.1 & \underline{67.5} & 93.2 & 65.0 & 70.6 & 62.6 & 67.8 & 47.4 & 67.1 \\
Focus & 65.7 & 55.3 & 54.6 & 73.5 & 62.3 & 63.9 & 58.0 & 68.0 & 55.8 & 61.9 \\ 
Verb           & 64.0 & 63.0 & 67.0 & 98.0 & \textbf{72.0} & \textbf{88.0} & \textbf{75.0} & 64.0 & 57.0 & \underline{72.2} \\
\rowcolor{gray!10}
\textit{GPT-4.1} & \textit{91.7}    & \textit{83.4} & \textit{62.8} & \textit{99.1}    & \textit{73.3}    & \textit{89.0}    & \textit{85.1}    & \textit{49.1}   & \textit{71.5} & \textit{78.3} \\
\midrule
\multicolumn{11}{c}{\textit{Retrieval-Free Supervised Methods}} \\
\midrule
MM           & 74.8 & 59.7 & 48.4 & 58.0 & 67.8 & 60.3 & 52.0 & 56.8 & 50.3 & 58.7 \\
CCS          & 61.0 & 48.8 & 46.3 & 53.6 & 56.8 & 51.2 & 51.0 & 55.0 & 53.2 & 53.0 \\
TAD & 82.0 & 62.4 & 59.2 & 71.5 & 63.8 & 62.6 & 52.4 & 69.3 & 51.2 & 63.8 \\
SATRMD & 78.7 & 63.3 & 54.2 & 81.5 & 69.0 & 69.8 & 58.0 & 60.8 & 52.8 & 65.3 \\
MIND & 85.9 & 61.6 & 59.6 & 96.2 & 68.3 & 60.1 & 63.2 & 61.3 & 45.0 & 66.8 \\
Sheeps & \textbf{87.2} & 56.8 & 60.2 & \underline{98.5} & {70.7} & {81.0} & {67.9} & \underline{69.6} & 52.1 & {71.6} \\
SAPLMA & 85.8 & \underline{64.3} & 61.1 & 85.4 & 68.0 & 68.6 & 58.0 & 63.8 & 60.9 & 68.4 \\
\midrule
\textbf{INTRA} & \underline{86.5} & 56.2 & 60.1 & \textbf{99.0} & 69.1 & \underline{84.1} & \underline{72.4} & 69.3 & 58.7 & \textbf{72.8} \\
\bottomrule
\end{tabular}}
\end{table*}

\begin{table*}[h!]
\centering
\caption{Performance of claim verification methods for \textbf{Ministral-8B}  in the proposed evidence-free setting, measured by PR-AUC$\uparrow$ (multiplied by 100) across nine datasets. \textbf{Bold} values indicate the best result per dataset, the second best is \underline{underlined}, and \textit{Avg} the mean score across all datasets.}
\label{tab:main_prauc_ministral8b}
\resizebox{\textwidth}{!}{
\begin{tabular}{l|cccccccccc}
\toprule
\multirow{2}{*}{\textbf{Method}}
& \textit{\scriptsize long-tail}
& \textit{\scriptsize human-made}
& \textit{\scriptsize cross-model}
& \multicolumn{4}{c}{\textit{\scriptsize rule-generated}}
& \textit{\scriptsize long-form}
& \textit{\scriptsize multilingual}
& \multirow{2}{*}{\textbf{Avg}}
\\
\cmidrule(lr){5-8}
 & \textbf{PopQA} & \textbf{AVeriTeC} & \textbf{UHead} & \textbf{Cities} & \textbf{CC} & \textbf{CMP} & \textbf{CF} & \textbf{WH} & \textbf{X-Fact} &  \\
 \midrule
\multicolumn{11}{c}{\textit{Retrieval-Based Methods}} \\  
 \midrule
 Verb + RAG & 87.0 & 85.0 & 47.0 & 96.0 & 67.0 & 81.0 & 67.0 & 31.0 & 67.0 & 69.6 \\
\midrule
\multicolumn{11}{c}{\textit{Retrieval-Free Unsupervised Methods}} \\
\midrule

SP & 87.4 & \textbf{78.1} & \underline{39.0} & 93.7 & 63.4 & 63.6 & 60.9 & 33.8 & 64.6 & 64.9 \\
PPL & 87.7 & 71.6 & 28.2 & 95.5 & \textbf{70.3} & 71.1 & 58.9 & 22.3 & 72.3 & 64.2 \\
MTE & 83.3 & 75.8 & 30.0 & 51.5 & 68.2 & 54.4 & 50.6 & 31.8 & \textbf{76.6} & 58.0 \\
Att. Score & 75.3 & 75.5 & 32.1 & 48.1 & 45.3 & 48.4 & 49.5 & 23.5 & 61.4 & 51.0 \\
RAUQ & 85.5 & 72.5 & 28.2 & 87.4 & 67.2 & 56.7 & 54.6 & 22.4 & 72.2 & 60.7 \\
CCP & 86.0 & 75.0 & \textbf{43.1} & 93.4 & 65.2 & 69.4 & 60.8 & 32.1 & 64.7 & 65.5 \\
Focus & 82.5 & 73.9 & 28.2 & 66.4 & 61.6 & 61.2 & 55.8 & 30.9 & 69.7 & 58.9 \\
Verb           & 82.0 & 76.0 & 36.0 & 97.0 & 65.0 & \underline{83.0} & \textbf{68.0} & 25.0 & 69.0 & 66.5 \\
\rowcolor{gray!10}
\textit{GPT-4.1} & \textit{91.7}    & \textit{83.4} & \textit{62.8} & \textit{99.1}    & \textit{73.3}    & \textit{89.0}    & \textit{85.1}    & \textit{49.1}   & \textit{71.5} & \textit{78.3} \\
\midrule
\multicolumn{11}{c}{\textit{Retrieval-Free Supervised Methods}} \\
\midrule
MM           & 89.5 & 77.5 & 25.7 & 58.5 & 67.0 & 60.9 & 51.2 & 19.5 & 63.8 & 57.1 \\
CCS          & 82.0 & 70.6 & 23.8 & 51.6 & 54.8 & 51.4 & 50.6 & 16.7 & 65.1 & 51.8 \\
TAD & 92.8 & 76.3 & 31.2 & 70.5 & 62.5 & 62.6 & 51.9 & 33.5 & 67.3 & 60.9 \\
SATRMD & 91.2 & 77.8 & 28.6 & 81.1 & 67.8 & 67.8 & 55.9 & 24.8 & 67.4 & 62.5 \\
MIND & \underline{94.6} & 77.6 & 33.0 & 97.0 & 68.1 & 60.9 & 59.3 & 26.5 & 62.6 & 64.4 \\
Sheeps & \textbf{95.0} & 72.4 & 31.8 & \underline{98.1} & \underline{69.8} & {78.4} & {62.2} & \underline{34.3} & 66.1 & \underline{67.6} \\
SAPLMA & 94.3 & \underline{77.9} & 35.9 & 83.8 & 65.5 & 67.1 & 55.9 & 26.5 & \underline{73.2} & 64.5 \\
\midrule
\textbf{INTRA} & 94.5 & 71.3 & 34.0 & \textbf{98.9} & 68.0 & \textbf{84.0} & \underline{67.1} & \textbf{34.9} & 70.9 & \textbf{69.3} \\

\bottomrule
\end{tabular}}
\end{table*}

\clearpage
\begin{table*}[h!]
\centering
\caption{Performance of claim verification methods for \textbf{Phi-4-mini} in the proposed evidence-free setting, measured by ROC-AUC$\uparrow$ across nine datasets. \textbf{Bold} values indicate the best result per dataset, the second best is \underline{underlined}, and \textit{Avg} the mean score across all datasets. 
}
\label{tab:main_rocauc_phi4}
\resizebox{\textwidth}{!}{
\begin{tabular}{l|cccccccccc}
\toprule
\multirow{2}{*}{\textbf{Method}}
& \textit{\scriptsize long-tail}
& \textit{\scriptsize human-made}
& \textit{\scriptsize cross-model}
& \multicolumn{4}{c}{\textit{\scriptsize rule-generated}}
& \textit{\scriptsize long-form}
& \textit{\scriptsize multilingual}
& \multirow{2}{*}{\textbf{Avg}}
\\
\cmidrule(lr){5-8}
 & \textbf{PopQA} & \textbf{AVeriTeC} & \textbf{UHead} & \textbf{Cities} & \textbf{CC} & \textbf{CMP} & \textbf{CF} & \textbf{WH} & \textbf{X-Fact} &  \\
\midrule
\multicolumn{11}{c}{\textit{Retrieval-Based Methods}} \\  
 \midrule
 Verb + RAG & 78.0 & 72.0 & 71.0 & 95.0 & 73.0 & 87.0 & 74.0 & 66.0 & 73.0 & 76.5 \\

\midrule
\multicolumn{11}{c}{\textit{Retrieval-Free Unsupervised Methods}} \\  
 \midrule
SP & 71.5 & \textbf{64.0} & \textbf{66.2} & 92.2 & 63.8 & 65.0 & 61.6 & 67.4 & 45.8 & 66.4 \\
Perplexity & 72.2 & 55.0 & 60.6 & 91.6 & 70.7 & 71.1 & 61.9 & 61.0 & 55.0 & 66.6 \\
MTE & 64.4 & 61.1 & 61.1 & 54.1 & 70.4 & 56.6 & 50.9 & 69.1 & 59.4 & 60.8 \\
Att. Score & 50.6 & 59.5 & 59.1 & 40.7 & 42.1 & 48.9 & 49.5 & 59.0 & 43.6 & 50.3 \\
RAUQ & 65.9 & 59.5 & 57.9 & 81.2 & 65.1 & 54.9 & 53.7 & 60.5 & 55.4 & 61.6 \\
CCP & 47.0 & 54.6 & 58.0 & 43.5 & 40.8 & 48.6 & 49.3 & 53.5 & 42.5 & 48.6 \\
Focus & 49.7 & 49.9 & 49.0 & 50.0 & 51.2 & 50.2 & 50.0 & 49.8 & 52.0 & 50.2 \\
Verb & 63.0 & 53.0 & 60.0 & 96.0 & 72.0 & \textbf{88.0} & \textbf{75.0} & 65.0 & 51.0 & 69.1 \\
\rowcolor{gray!10}
\textit{GPT-4.1} & \textit{84.7} & \textit{76.7} & \textit{87.0} & \textit{99.5} & 8\textit{0.3} & \textit{92.3} & \textit{90.8} & \textit{81.8} & \textit{62.8} & \textit{84.0} \\
\midrule
\multicolumn{11}{c}{\textit{Supervised Methods}} \\
\midrule
MM & 81.7 & 60.8 & 61.3 & \textbf{99.5} & \underline{79.3} & \underline{86.2} & 68.9 & \textbf{70.0} & 51.7 & \textbf{73.3} \\
CCS & 82.3 & \underline{61.4} & 59.5 & \underline{99.3} & \textbf{79.5} & 85.5 & \underline{69.6} & 67.9 & 52.6 & \underline{73.1} \\
TAD & 81.4 & 56.7 & 57.2 & 84.4 & 63.1 & 60.9 & 53.6 & \textbf{70.0} & 54.8 & 64.7 \\
SATRMD & 73.8 & 60.5 & 58.6 & 89.0 & 69.0 & 66.5 & 59.0 & 58.6 & 48.8 & 64.9 \\
MIND & 83.9 & 60.1 & \underline{64.6} & 95.9 & 72.1 & 73.7 & 57.9 & 69.2 & 52.8 & 70.0 \\
Sheeps & \textbf{84.9} & 56.7 & 62.6 & 90.2 & 67.6 & 73.0 & 60.1 & 69.8 & \textbf{59.4} & 69.4 \\
SAPLMA & 82.1 & 48.0 & 61.7 & 78.9 & 68.8 & 67.3 & 58.4 & 66.3 & 56.8 & 65.4 \\
\midrule
\textbf{INTRA} & \underline{84.0} & 55.8 & 61.9 & 91.7 & 66.1 & 76.4 & 61.0 & \underline{69.9} & \underline{58.3} & 69.5 \\
\bottomrule
\end{tabular}}
\end{table*}
\begin{table*}[h!]
\centering
\small
\caption{Performance of claim verification methods for \textbf{Phi-4-mini}  in the proposed evidence-free setting, measured by PR-AUC$\uparrow$ (multiplied by 100) across nine datasets. \textbf{Bold} values indicate the best result per dataset, the second best is \underline{underlined}, and \textit{Avg} the mean score across all datasets. 
}
\label{tab:main_prauc_phi4}
\resizebox{\textwidth}{!}{
\begin{tabular}{l|cccccccccc}
\toprule
\multirow{2}{*}{\textbf{Method}}
& \textit{\scriptsize long-tail}
& \textit{\scriptsize human-made}
& \textit{\scriptsize cross-model}
& \multicolumn{4}{c}{\textit{\scriptsize rule-generated}}
& \textit{\scriptsize long-form}
& \textit{\scriptsize multilingual}
& \multirow{2}{*}{\textbf{Avg}}
\\
\cmidrule(lr){5-8}
 & \textbf{PopQA} & \textbf{AVeriTeC} & \textbf{UHead} & \textbf{Cities} & \textbf{CC} & \textbf{CMP} & \textbf{CF} & \textbf{WH} & \textbf{X-Fact} &  \\
\midrule
\multicolumn{11}{c}{\textit{Retrieval-Based Methods}} \\  
 \midrule
 Verb + RAG & 88.0 & 81.0 & 41.0 & 91.0 & 66.0 & 82.0 & 66.0 & 24.0 & 66.0 & 67.1 \\

\midrule
\multicolumn{11}{c}{\textit{Retrieval-Free Unsupervised Methods}} \\  
 \midrule
SP & 86.0 & \underline{78.2} & \textbf{38.6} & 87.8 & 62.4 & 61.4 & 59.0 & 28.6 & 63.0 & 62.8 \\
Perplexity & 87.7 & 73.4 & 31.2 & 90.9 & 68.5 & 69.2 & 58.7 & 21.7 & 66.2 & 63.1 \\
MTE & 82.0 & 74.5 & 30.5 & 52.3 & 68.0 & 54.8 & 50.5 & 29.3 & 67.6 & 56.6 \\
Att. Score & 75.5 & 75.5 & 31.4 & 45.3 & 43.9 & 48.9 & 49.7 & 23.3 & 61.7 & 50.6 \\
RAUQ & 84.4 & 77.2 & 28.8 & 79.4 & 63.5 & 53.1 & 52.6 & 21.5 & 67.2 & 58.6 \\
CCP & 73.1 & 74.0 & 29.4 & 45.2 & 42.9 & 48.5 & 49.5 & 18.9 & 61.1 & 49.2 \\
Focus & 74.8 & 68.4 & 15.9 & 75.0 & 59.1 & 62.6 & 50.9 & 18.1 & \textbf{71.0} & 55.1 \\
Verb           & 81.0 & 71.0 & 31.0 & 92.0 & 65.0 & 83.0 & \textbf{67.0} & 25.0 & 66.0 & 64.5 \\
\rowcolor{gray!10}
\textit{GPT-4.1} & \textit{91.7}    & \textit{83.4} & \textit{62.8} & \textit{99.1}    & \textit{73.3}    & \textit{89.0}    & \textit{85.1}    & \textit{49.1}   & \textit{71.5} & \textit{78.3} \\
\midrule
\multicolumn{11}{c}{\textit{Retrieval-Free Supervised Methods}} \\
\midrule
MM & 92.9 & 77.5 & 32.3 & \textbf{99.5} & \underline{76.6} & \textbf{85.1} & 64.1 & 26.6 & 63.3 & \underline{68.7} \\
CCS & 93.1 & \textbf{78.3} & 30.1 & \underline{99.4} & \textbf{78.6} & \underline{84.5} & \underline{66.0} & 25.8 & 64.3 & \textbf{68.9} \\
TAD & 92.2 & 71.9 & 32.0 & 80.6 & 61.3 & 57.8 & 52.5 & \textbf{33.4} & 69.5 & 61.2 \\
SATRMD & 88.1 & 76.2 & 29.0 & 87.2 & 66.7 & 63.0 & 56.4 & 21.3 & 63.0 & 61.2 \\
MIND & \underline{93.6} & 75.9 & 34.3 & 94.5 & 70.0 & 68.7 & 55.4 & 30.3 & 66.1 & 65.4 \\
Sheeps & \textbf{93.8} & 73.9 & 33.9 & 87.2 & 66.0 & 68.0 & 56.3 & 32.1 & \underline{70.6} & 64.6 \\
SAPLMA & 93.1 & 68.9 & \underline{35.0} & 78.6 & 66.4 & 64.7 & 56.2 & 30.8 & 69.8 & 62.6 \\
\midrule
\textbf{INTRA} & 93.3 & 72.7 & 33.7 & 87.5 & 64.5 & 71.7 & 56.8 & \underline{32.2} & 69.1 & 64.6 \\
\bottomrule
\end{tabular}}
\end{table*}

\end{document}